%% file: main.tex
\documentclass[journal]{IEEEtran}
\usepackage{nameref}

\usepackage[utf8]{inputenc} 
\usepackage[T1]{fontenc}    

\usepackage[backend=biber,abbreviate=true,doi=false,url=false,isbn=false,eprint=false,style=ieee]{biblatex}
\usepackage{hyperref}       
\usepackage{url}            
\usepackage{booktabs}       
\usepackage{amsfonts}       
\usepackage{amsmath} 
\usepackage{amssymb}
\usepackage{nicefrac}       
\usepackage[super]{nth}
\usepackage{siunitx}

\usepackage{array}

\usepackage{graphicx} 
\graphicspath{{./figures/}}

\begin{document}

\title{State-Space Constraints Can Improve the Generalisation of the Differentiable Neural Computer to Input Sequences With Unseen Length}

\author{Patrick~Ofner and Roman~Kern
\thanks{P. Ofner was with the Area Knowledge Discovery, Know-Center Research GmbH, Graz, Austria. He is currently with the Department of Microsystems Engineering, University of Freiburg, Freiburg, Germany (e-mail: patrick@ofner.science).}
\thanks{R. Kern is with the Institute of Machine Learning and Neural Computation, Graz University of Technology, Graz, Austria, and the Know-Center Research GmbH, Graz, Austria (e-mail: rkern@tugraz.at).}}

\maketitle

\begin{abstract}
Memory-augmented neural networks (MANNs) can perform algorithmic tasks such as sorting. However, they often fail to generalise to input sequence lengths not encountered during training. We introduce two approaches that constrain the state space of the MANN's controller network: \emph{state compression} and \emph{state regularisation}. We empirically demonstrated that both approaches can improve generalisation to input sequences of out-of-distribution lengths for a specific type of MANN: the differentiable neural computer (DNC). The constrained DNC could process input sequences that were up to 2.3 times longer than those processed by an unconstrained baseline controller network. Notably, the applied constraints enabled the extension of the DNC's memory matrix without the need for retraining and thus allowed the processing of input sequences that were 10.4 times longer. However, the improvements were not consistent across all tested algorithmic tasks. Interestingly, solutions that performed better often had a highly structured state space, characterised by state trajectories exhibiting increased curvature and loop-like patterns. Our experimental work demonstrates that state-space constraints can enable the training of a DNC using shorter input sequences, thereby saving computational resources and facilitating training when acquiring long sequences is costly.


\end{abstract}

\begin{IEEEkeywords}
differentiable neural computer, memory-augmented neural network, generalisation, state-space, algorithmic tasks
\end{IEEEkeywords}

\section{Introduction}
\label{sec:introduction}

\IEEEPARstart{T}{}he research on artificial neural networks (ANNs) for solving algorithmic problems has gained momentum over the past decade. Compared to symbolic approaches, neural networks can be beneficial in situations where symbolic methods struggle with missing, noisy, or inconsistent data, as they can effectively interpolate and abstract complex input data \cite{Numeroso2023}. Although ANNs may struggle with generalisation when presented with out-of-distribution data, they can provide a fast and sufficiently good heuristic suited for real-world applications \cite{BeurerKellner2022}. Additionally, ANN-based techniques can indicate to users which parts of the input to concentrate on during complex tasks, such as proving mathematical theorems \cite{Davies2021}.

Modern large language models (LLMs) based on transformers \cite{Brown2020} can solve algorithmic problems \cite{Anil2022,Meijer2024}, but require substantial computational resources and large datasets for training. Recent large reasoning models \cite{OpenAI2024,DeepSeekAI2025,Anthropic2025}
further improve reasoning performance through extended inference-time computation 
and self-reflection mechanisms \cite{Wei2022,Wang2023}. However, they demand even greater computational budgets, particularly during inference, while still exhibiting accuracy collapses beyond certain complexity thresholds \cite{Shojaee2025}. In contrast, neural program synthesis approaches, which are based on the assembly of primitive processing units, are efficiently trainable and can generalise, but they often require supervision with execution traces that impede end-to-end learning \cite{Reed2016,Kurach2016,Cai2017,Li2017,Yan2020}. Similarly, neural reasoning with graph neural networks suffers from the same limitation \cite{Velickovic2020,Ibarz2022}. Thus, neural architectures that are efficiently end-to-end trainable are desirable, particularly when computational resources are constrained or acquiring long training sequences is costly.

Neural architectures based on recurrent neural networks (RNNs) \cite{Zaremba2014,Kalchbrenner2016,Santoro2018,Goyal2019}, transformer \cite{Csordas2022,Reed2022,Petruzzellis2024}, or other neural architectures \cite{Vinyals2015,Kaiser2016,Huynh2020,Madsen2020} have been shown to be able to solve algorithmic tasks by neural program induction from training samples in an end-to-end manner.


One particularly promising idea is to equip an ANN with an \emph{external memory} compatible with gradient descent-based training algorithms to solve algorithmic problems \cite{Graves2014,Sukhbaatar2015}. Such networks are often classified as memory-augmented neural networks (MANNs) and are situated above RNN, CNN, or Transformer in the Chomsky hierarchy \cite{Deletang2023}. MANNs can be classified according to their memory access scheme. Prominent examples include stack \cite{Joulin2015,Suzgun2019,Stogin2024} or random access memory schemes \cite{Graves2014,Graves2016,Kurach2016,Paassen2022}, an overview is given in~\cite{Ma2020}. A well-known representative of the latter access scheme is the \emph{Neural Turing Machine} NTM\cite{Graves2014} or its successor the \emph{Differentiable Neural Computer} (DNC) \cite{Graves2016}. The capabilities of the DNC have been demonstrated in, for example, question answering from short stories, route finding in the London Underground (i.e.\ graph traversal), or mastering a block puzzle game \cite{Graves2016}. Both NTM and DNC comprise an ANN-based \emph{controller} and a \emph{memory matrix}. The controller network learns via gradient descent to (1) compute the output based on the input and memory content, and (2) read and write to the memory matrix via attention mechanisms \cite{Bahdanau2015}. The controller network could be any type of ANN, such as a feed-forward ANN or a Long short-term memory (LSTM) \cite{Hochreiter1997}.

A central performance aspect for ANNs addressing algorithmic tasks is their ability to generalise to out-of-distribution samples, which refers to situations where the distribution of test data differs from that of the training data. In algorithmic tasks, such differences may be evident in the set of input symbols or the length of input sequences. The latter is particularly important in practice, as the training time for MANNs increases with longer input sequences. Effective input length generalisation would allow MANNs to be trained on sequences shorter than those anticipated during inference, thus saving computation time. Furthermore, mechanisms that facilitate input length generalisation may offer valuable insights into the data processing mechanisms of neural networks.

MANN architectures, such as the NTM or DNC, demonstrate only a limited ability to generalise to longer input sequences. While there has been some progress in performance improvements in general (c.f.\ \cite{Rae2016, Taguchi2018, Franke2018, Csordas2019, Park2020}), specific research on enhancing input length generalisation for NTM and DNC remains scarce. Le et al.\ \cite{Le2020} proposed a method to dynamically load controller weights through attention mechanisms from memory rather than relying on fixed weights. This improved input-length generalisation for tasks like copy tasks. This concept was further developed by enabling the controller weights to represent shareable and modular programs \cite{Le2020a}. Yang et al.\ \cite{Yang2017} utilised group theory to formalise memory access as group actions, demonstrating that a differentiable group (i.e.,\ Lie group) allows for input-length generalisation to sequences that are double the length of those encountered during training for tasks like sorting and copying. Moreover, it was also shown that sparse memory access can enhance generalisation \cite{Rae2016}.

In this study, we investigated a novel approach to improve the DNC's input length generalisation. We assumed that the neural state of the DNC's controller network evolves along a trajectory with minimal curvature and eventually enters a region of the state-space that was not encountered during training. This could cause the DNC to exhibit unexpected behaviour with unseen lengths of input sequences. We hypothesised that imposing constraints on the state-space would result in potentially more structured state trajectories with greater curvature, thereby improving input length generalisation. The intuition behind this is that such trajectories could form structures, such as loops. Returning to states via loops could enhance input length generalisation since states would remain within the region encountered during training.

To constrain the state space, we (1) \emph{compressed} the state-space of the DNC's controller and/or (2) \emph{regularised} the state-space so that subsequent controller states were close to previously visited ones. Compression was realised through an architectural adaptation, whereas regularisation was implemented with an additional loss term. Furthermore, we investigated whether expanding the memory matrix, a characteristic of MANNs, would improve input length generalisation. Our methods were tested on a set of basic algorithmic tasks with input sequence lengths that were not seen during training. We analysed the curvature of the state trajectories and visually examined their structure in state space.












\section{Background}
\label{sec:background}
\begin{figure}
    \centering
    \includegraphics[width=1\linewidth]{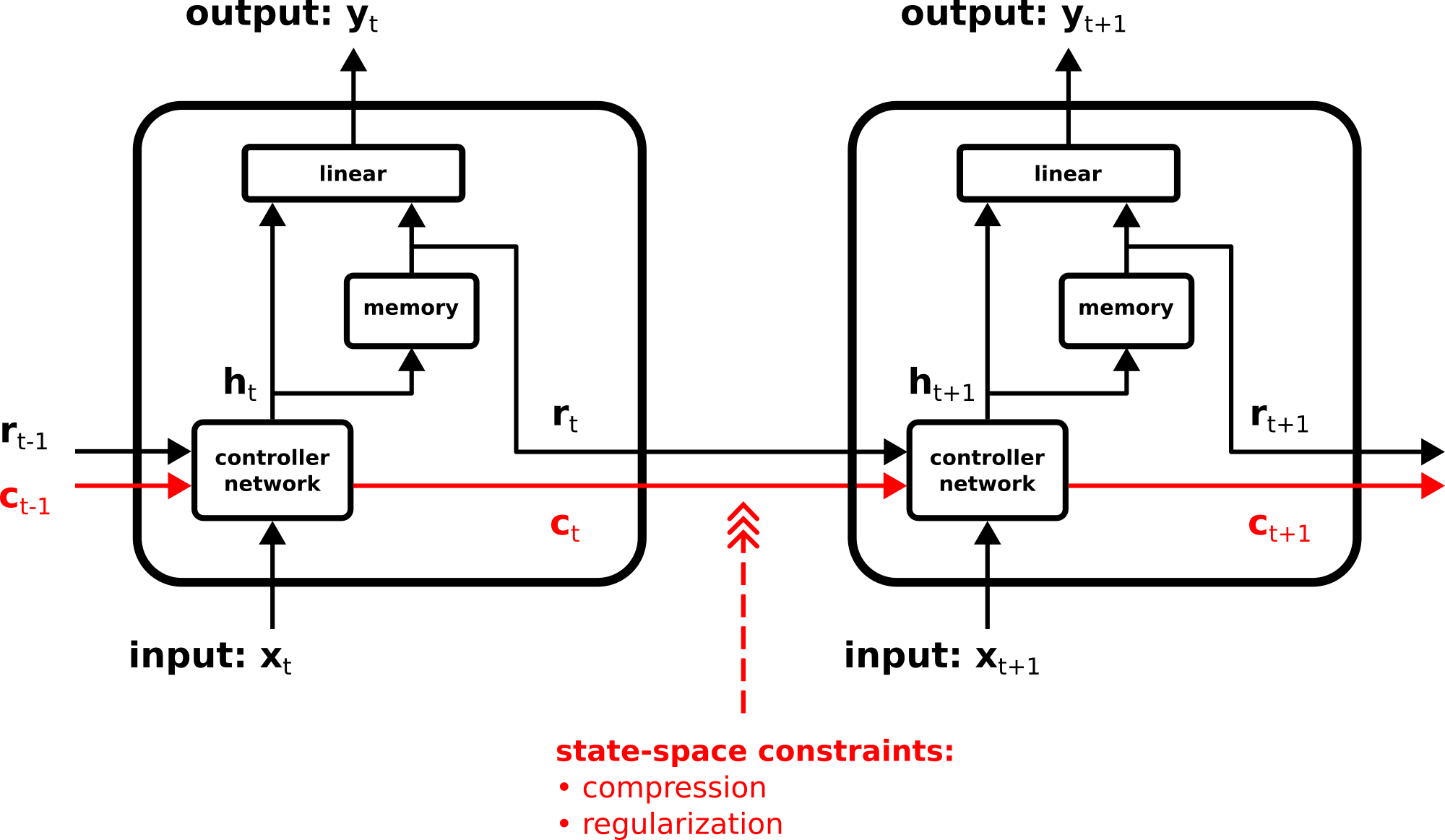}
    \caption{Architecture of the Differentiable Neural Computer (DNC). We imposed constraints on the cell state $\mathbf{c}_t$ (through compression and/or regularisation) to enhance generalisation to longer input sequences.}
    \label{fig:overview}
\end{figure}

The DNC\cite{Graves2016} is an RNN and comprises a controller and a dedicated memory matrix, which the controller accesses via attention mechanisms (see Fig.~\ref{fig:overview}).

\paragraph{DNC Memory}
The memory $\mathbf{M}$ is organised as an $N \times W$ matrix, where the $N$ rows are memory slots with a width of $W$.
An attention mechanism \cite{Bahdanau2015} with $\mathbf{w}_t^r$ and $\mathbf{w}_t^w \in \mathbb{R}^N$ being the read and write weight vectors, respectively, is used to access the memory.

Data is read from the memory and stored in a read vector $\mathbf{r}_t \in \mathbb{R}^W$ as:
\[\mathbf{r}_t = \mathbf{M}_t^\top \mathbf{w}_t^r\]
where $t$ is the time index. Writing to memory is implemented via a combined erase and write operation:
\[\mathbf{M}_t = \mathbf{M}_{t-1} \odot (\mathbf{E} - \mathbf{w}_t^w \mathbf{e}_t^\top) + \mathbf{w}_t^w \mathbf{v}_t^\top\]
where $\odot$ denotes the element-wise matrix multiplication, $\mathbf{E}$ is an all-ones matrix of size $N \times W$, $\mathbf{e}_t \in [0,1]^W$ is the erase vector and $\mathbf{v}_t \in \mathbb{R}^W$ is the write vector.

The read and write weight vectors $\mathbf{w}_t^r$ and $\mathbf{w}_t^w$ are computed by read and write heads, respectively. The read and write heads support in total three memory attention mechanisms: (1) \emph{content-based writing and reading} via a key/value mechanism, where the controller generates the keys, and weight vectors are based on a similarity measure between a key and each memory slot; (2) writing via \emph{dynamic memory allocation}; (3) \emph{sequential reading} from memory slots in the forward or backward direction of previously written sequences. 
The read and write heads are controlled by the interface vector $\mathbf{\xi}_t$, which the controller emits. $\mathbf{\xi}_t$ configures the memory attention mechanisms and contains the data to be written into the memory. The read and write heads compute then $\mathbf{w}_t^r$ and $\mathbf{w}_t^w $, respectively, based on $\mathbf{\xi}_t$ and the memory content $\mathbf{M}$ (see \cite{Graves2016} for details).

Multiple read heads can be employed, and we use for the remainder of the paper an additional read head index $i$ for the read weight vector $\mathbf{w}_t^{r,i}$ and read vector $\mathbf{r}_t^i$.

\paragraph{DNC Controller}
The controller reads the input, accesses the memory, and produces the output of the DNC. The central part of the controller is the \emph{controller network}.
In addition, the controller includes linear transformations, represented by matrices $\mathbf{W}_y, \mathbf{W}_r$, and $\mathbf{W}_{\xi}$, which produce the DNC output $\mathbf{y}_t$ and the interface vector $\mathbf{\xi}_t$. Formally, the controller receives an input vector $\mathbf{x}_t \in \mathbb{R}^X$ and read vectors $\mathbf{r}_{t-1}^i \in \mathbb{R}^W$ of the $R$ read heads from the previous time step and subsequently computes the DNC output vector $\mathbf{y}_t \in \mathbb{R}^Y$ and interface vector $\mathbf{\xi}_t$.
For notational convenience, we concatenate the input vector and memory readouts:
\[\mathbf{\chi}_t = [\mathbf{x}_t; \mathbf{r}_{t-1}^1; \ldots; \mathbf{r}_{t-1}^R]\]
with $\mathbf{\chi}_t \in \mathbb{R}^{X + RW}$.
The controller network processes the input $\mathbf{\chi}_t$ and produces the controller network output $\mathbf{h}_t \in \mathbb{R}^H$ as $\mathbf{h}_t = ANN([\mathbf{\chi}_1; \dots;\mathbf{\chi}_t]; \theta)$, where $ANN$ is a neural network (possibly recurrent) and $\theta$ are the network weights. The controller network output and the read vectors are then linearly transformed to yield the DNC output:
\[\mathbf{y}_t = \mathbf{W}_y \mathbf{h}_t + \mathbf{W}_r[\mathbf{r}_t^1;\ldots;\mathbf{r}_t^R] + \mathbf{b}_y\]
where $\mathbf{W}_y$ is the output matrix of size $Y \times H$, $\mathbf{W}_r$ is the readout matrix of size $Y \times RW$, and $\mathbf{b}_y$ is a bias vector.

The interface vector $\mathbf{\xi}_t$ is computed as:
\[\mathbf{\xi}_t = \mathbf{W}_{\xi} \mathbf{h}_t + \mathbf{b}_{\xi}\]
where $\mathbf{W}_{\xi}$ is of size $(WR+3W+5R+3) \times H$ and called interface matrix, and $\mathbf{b}_{\xi}$ is the corresponding bias vector. $\mathbf{\xi}_t$ is interpreted by the read and write heads, and thus controls the memory attention mechanisms and the data written to and read from memory \cite{Graves2016}.

$\theta$, $\mathbf{W}_{\{y,r,\xi\}}$ and $\mathbf{b}_{\{y,\xi\}}$ are usually found by an optimization algorithm.



\section{Approaches}
\label{sec:approaches}

The DNC controller is typically implemented using an RNN (such as an LSTM \cite{Hochreiter1997}) and thus possesses \emph{state variables}. We propose two approaches to facilitate generalisation concerning the length of the input sequence: \emph{compression} and \emph{regularisation} of the controller's state space (see Fig.~\ref{fig:overview}). We evaluated these approaches using a set of simple algorithmic tasks, comparing them to both stateful and stateless controllers as baselines. Below, we outline the stateful baseline (A), the stateless baseline (B), the architectural change underlying both our methods (C), and our approaches to state-space constraints (D and E).

\subsection{Stateful Baseline}
\label{sec:baseline controller}
An LSTM controller network \cite{Hochreiter1997} was employed as a baseline. As an LSTM is stateful, the controller could compute the DNC output by considering its current state in addition to the input and the memory matrix.
The LSTM was defined as:
\begin{align*}
\mathbf{i}_t &= \sigma (\mathbf{W}_i [\mathbf{\chi}_t; \mathbf{h}_{t-1}] + \mathbf{b}_i) \\
\mathbf{f}_t &= \sigma (\mathbf{W}_f [\mathbf{\chi}_t; \mathbf{h}_{t-1}] + \mathbf{b}_f) \\
\mathbf{o}_t &= \sigma (\mathbf{W}_o [\mathbf{\chi}_t; \mathbf{h}_{t-1}] + \mathbf{b}_o)] \\
\mathbf{c}_t &= \mathbf{f}_t \mathbf{c}_{t-1} + \mathbf{i}_t \tanh ( \mathbf{W}_c [\mathbf{\chi}_t; \mathbf{h}_{t-1}] + \mathbf{b}_c) \\
\mathbf{h}_t &= \mathbf{o}_t \tanh (\mathbf{c}_t)
\end{align*}
where $\sigma$ is the logistic sigmoid function, and $\mathbf{i}_t$, $\mathbf{f}_t$, $\mathbf{o}_t$, $\mathbf{c}_t$, and $\mathbf{h}_t \in \mathbb{R}^H$
are the input gate, forget gate, output gate, cell state,  and hidden state vectors, respectively, at time $t$.
$\mathbf{W}_i$, $\mathbf{W}_f$, $\mathbf{W}_o$, $\mathbf{W}_c \in \mathbb{R}^{H \times (X + RW)}$, and $\mathbf{b}_i$, $\mathbf{b}_f$, $\mathbf{b}_o$, $\mathbf{b}_c \in \mathbb{R}^H$
are trainable weight matrices and bias vectors, respectively, and are referred to as $\theta$ in this work.

\subsection{Stateless Baseline}
\label{sec:stateless controller}
A controller using a feedforward neural network (FFNN) served as a stateless baseline:
\[\mathbf{h}_t = FFNN_{\theta}(\mathbf{\chi}_t),\]
where $FFNN_{\theta}()$ is a fully connected neural network with potentially multiple hidden layers and parameterised by $\theta$. The stateless baseline can be considered an extreme constraint of the stateful baseline. This is because this type of controller has no internal memory and therefore no state space.

\subsection{Peephole LSTM}
\label{sec:peephole lstm}
We constrained the state space of a controller realised with an LSTM. For this purpose, we applied compression and regularisation to the LSTM cell state $\mathbf{c}_t$. However, we did not constrain the LSTM hidden state $\mathbf{h}_t$, since the interface vector $\mathbf{\xi}_t$ was derived from it, and applying constraints could have hindered memory access. To prevent cell state constraints from being evaded via the hidden state, we adapted the LSTM so that only the cell state $\mathbf{c}_t$, but not the hidden state $\mathbf{h}_t$, was propagated through time.
Thus, the gates $\mathbf{i}_t$, $\mathbf{f}_t$, and $\mathbf{o}_t$ had connections to $\mathbf{c}_t$ instead of $\mathbf{h}_t$:
\begin{align*}
\mathbf{i}_t &= \sigma (\mathbf{W}_i [\mathbf{\chi}_t; \mathbf{c}_{t-1}] + \mathbf{b}_i) \\
\mathbf{f}_t &= \sigma (\mathbf{W}_f [\mathbf{\chi}_t; \mathbf{c}_{t-1}] + \mathbf{b}_f) \\
\mathbf{o}_t &= \sigma (\mathbf{W}_o [\mathbf{\chi}_t; \mathbf{c}_{t-1}] + \mathbf{b}_o)] \\
\mathbf{c}_t &= \mathbf{f}_t \mathbf{c}_{t-1} + \mathbf{i}_t \tanh ( \mathbf{W}_c [\mathbf{\chi}_t; \mathbf{c}_{t-1}] + \mathbf{b}_c) \\
\mathbf{h}_t &= \mathbf{o}_t \tanh (\mathbf{c}_t)
\end{align*}
This adapted LSTM corresponds to a Peephole LSTM \cite{Gers2002,Sak2014} where the gates have direct access, i.e., peepholes, to the cell state. In contrast to a standard Peephole LSTM, the cell input activation vector $\tanh (.)$ used to compute $\mathbf{c}_t$ was also dependent on the previous cell state.

\subsection{State Compression}
\label{sec:state compression}
The range of the cell state vector $\mathbf{c}_t$ was limited, i.e., compressed, with the $\tanh$ function to the interval $(-1, 1)^H$. We redefined $\mathbf{c}_t$ and $\mathbf{h}_t$ of the previously defined Peephole LSTM as follows to implement state compression:
\begin{align*}
\mathbf{\widetilde{c}}_t &= \mathbf{f}_t \mathbf{c}_{t-1} + \mathbf{i}_t \tanh ( \mathbf{W}_c [\mathbf{\chi}_t; \mathbf{c}_{t-1}] + \mathbf{b}_c) \\
\mathbf{c}_t &= \tanh (\widetilde{\mathbf{c}_t}) \\
\mathbf{h}_t &= \mathbf{o}_t \tanh (\mathbf{\widetilde{c}}_t)
\end{align*}

\subsection{State Regularisation}
\label{sec:state regularization}
The state-space regularisation penalised distances between nearby states. We implemented state-space regularisation by adding a regularisation term to the loss function:
\[\mathcal{L} = \frac{1}{B} \sum_{i=1}^{B}(\lambda \mathcal{L}^i + (1-\lambda) \mathcal{L}_{state}^i),\]
where $B$ is the batch size, $\mathcal{L}^i$ is either the mean square error (mse) or cross-entropy loss of the $i$-th training sample, $\mathcal{L}_{state}^i$ is the additional state-space regularisation term, and $\lambda$ is a weighting coefficient.
First, the cosine similarity was calculated between every cell state pair of the $i$-th training sample:
\[\cos(\mathbf{c}_x^i, \mathbf{c}_y^i) = \frac{\langle \mathbf{c}_x^i, \mathbf{c}_y^i  \rangle}{\left|\left|\mathbf{c}_x^i\right|\right|
\left|\left|\mathbf{c}_y^i\right|\right|}\]
where $x$ and $y$ denote time indices during the processing of the training sample. Subsequently, the computed cosine similarities were sorted in descending order: Let $\pi_x^i$ and $\pi_y^i$ be two permutations of the cell state time indices for training sample $i$ such that $\cos(\mathbf{c}_{\pi^i_x(1)}^i, \mathbf{c}_{\pi^i_y(1)}^i) \geq \cos(\mathbf{c}_{\pi^i_x(2)}^i, \mathbf{c}_{\pi^i_y(2)}^i) \geq \cdots \geq \cos(\mathbf{c}_{\pi^i_x(n)}^i, \mathbf{c}_{\pi^i_y(n)}^i)$, where $\pi^i_x(t) \neq \pi^i_y(t)$. Finally, the $K$-largest cosine similarities (closest state pairs) were averaged and transformed into a loss:
\[\mathcal{L}_{state}^i = 1 - \frac{1}{K} \sum_{j=1}^K \cos(\mathbf{c}_{\pi_x(j)}^i, \mathbf{c}_{\pi_y(j)}^i)\]
Only the $K$-closest state pairs were considered in order to bring states within a neighbourhood closer together, but not states between neighbourhoods. This should result in an inductive bias towards state clusters and possibly the formation of loops.

\section{Experimental Design and Analyses}
\label{sec:experiments}
We evaluated how well our proposed generalisation approaches handle longer input sequences and analysed their impact on the controller's state-space. The following algorithmic tasks were employed: sort, copy, differentiation, shift, add, search, and logic evaluation. All tasks were tested with five different controller networks: as baselines, an LSTM and an FFNN; and as proposed approaches, a Peephole LSTM with state compression, a Peephole LSTM with state regularisation, and a Peephole LSTM with state compression \& state regularisation. We denote state compression and state regularisation as \emph{COMPR} and \emph{REG}, respectively, while their combination is referred to as \emph{COMPR\&REG}. The LSTM-based and FFNN-based controllers are termed \emph{STATEFUL-BASELINE} and \emph{STATELESS-BASELINE}, respectively.

Furthermore, we analysed whether a pre-trained DNC can be extended with a larger memory matrix without retraining and evaluated its performance on unseen input sequence lengths.

\subsection{Tasks}
\label{sec:tasks}
The algorithmic tasks employed in the experiments and their descriptions are provided in Table \ref{table:tasks}; their implementation details can be found in Section A
in supplementary materials. All tasks included an \emph{encoding phase} and a \emph{decoding phase}. In the encoding phase, the task-specific input was presented sequentially to the DNC. Subsequently, the computation result was read out sequentially in the decoding phase. The search task also included a \emph{search query phase} between the encoding and decoding phases (details in supplementary materials). Thus, for all tasks, the DNC received the complete task input before any output was generated, necessitating the DNC to utilise its memory. The total DNC input comprised the task-specific \emph{input} and a \emph{control channel} that separated the encoding and decoding phases. The total DNC output included the \emph{computation result} and, additionally, a \emph{signal channel} indicating the end of the output sequence (except for the logic task).

\begin{table*}[!h]
\centering
\caption{Algorithmic tasks for evaluating input length generalisation of the DNC. The examples illustrate the task input and expected output (auxiliary channels not shown). See Section A
in supplementary materials for task implementation details.}
\label{table:tasks}
	\begin{tabular}{c | c | c} 
	\hline
	 task & description & example\\
	 \hline
	 sort & sort a number sequence & $(2, 5, 2, 1, 3) \rightarrow (1, 2, 2, 3, 5)$ \\
	 copy & output a received number sequence & $(2, 5, 2, 1, 3) \rightarrow (2, 5, 2, 1, 3)$ \\
	 differentiation & output the absolute differences between successive numbers & $(2, 5, 2, 1, 3) \rightarrow (3, 3, 1, 2)$ \\
	 shift & circular shift a number sequence by its half & $(2, 5, 2, 1, 3) \rightarrow (1, 3, 2, 5, 2)$ \\
	 add & element-wise sum up two binary sequences & $((0,1), (0,0), (1,1), (1,0), (1,0)) \rightarrow (1, 0, 2, 1, 1)$ \\
	 search & output the relative positions of a queried numeral in a sequence & data: $(2, 5, 2, 1, 2)$, query: $(2)$ $\rightarrow (0, 0.5, 1)$ \\
	 logic evaluation & evaluate a propositional logic expression & $((\neg (\mathsf{T} \land \mathsf{F}) \to \mathsf{T}) \lor \mathsf{F}) \rightarrow \mathsf{T}$\\
	 \hline
	\end{tabular}
\end{table*}

\subsection{Network Implementation and Training}
\label{sec:implementation}
The LSTM and Peephole LSTM networks had 128 units (no stacking applied). The FFNN had three layers of 128 units each with $\tanh$ activation for the first two layers. All controllers had output size $H=128$. In the regularisation approach, the weighting coefficient of the loss function was set to $\lambda=0.9$, and cosine similarities were averaged over the closest $K=5$ state pairs. These settings were identified in preliminary experiments but were not systematically optimised. The neural networks were trained using the ADAM optimiser \cite{Kingma2015} to minimise the mean squared error or cross-entropy (for the logic evaluation task only) of the DNC's output in the decoding phase. The input sequences used for training ranged in length from $L_{in} = 5$ to 15. We trained 10 DNC instances, with different weight initialisations, for each task and generalisation approach. Details regarding network training can be found in Section B
in the supplementary materials. To implement our approaches and experiments, we extended the original DNC available on GitHub\footnote{\url{https://github.com/deepmind/dnc}}. We utilised Python 3.7 (Python Software Foundation) and Tensorflow 1.15 (Google LLC). Our extended DNC, the experimental framework, and the data are available on Zenodo\footnote{\url{https://doi.org/10.5281/zenodo.5534734}}.

\subsection{Performance Analysis}
\label{sec:performance evaluations}
Performance was measured using \emph{hit rate} for the search task and \emph{classification accuracy} for all others.
Classification accuracy was computed by comparing the DNC's computation result channel(s) with target values, averaged over the decoding phase. For the sort, copy, differentiation, shift, and add tasks, we disregarded the signal channel indicating output termination, as output length was determined by input length. For the search task, the DNC generated a sequence of position indices with termination indicated by a signal channel value $> 0.8$; hit rate was computed as the proportion of correct indices.

\paragraph{Input Sequence Length Generalisation}
We evaluated the performance of a DNC instance using out-of-distribution input sequences ranging in length from $L_{in} = 2$ to 100. We generated 10 test batches, each comprising $B=64$ samples. This gave us 640 test samples for each sequence length, which we then averaged to obtain the classification accuracy or hit rate. We refer to this evaluation procedure as a \emph{trial}. Ultimately, we repeated this procedure for every trained DNC instance. Thus, for each task and generalisation approach, we conducted 10 trials.

\paragraph{Memory Extension}
The memory capacity of a DNC can limit the length of sequences that can be processed. However, increasing the size of the memory matrix results in longer training times. To optimise computational resources, it may be preferable to increase the memory \emph{after} training the DNC. We investigated whether our generalisation approaches facilitate such a memory extension. To this end, we replaced the memory matrix $\mathbf{M}$ of a \emph{pre-trained} DNC with a 10x larger matrix ($N=500$) and evaluated the performance for input sequence lengths ranging from $L_{in}=2$ to 1000. To manage the increased computational demand during inference, we assessed performance using 3 batches instead of 10. Additionally, rather than assessing performance at each sequence length, we assessed it at progressively larger intervals.

\subsection{State-Space Analysis}
\label{sec:state-space analysis}
Each trained DNC instance was presented with one batch ($B=64$) of input sequences of length $L_{in} = 30$, which was not seen during training. This enabled us to observe the controller state-space when processing out-of-distribution samples. To allow for a fair comparison of our proposed regularisation approaches, we obtained the cell state output before applying any compression function ($\tanh$), i.e., we used $\widetilde{\mathbf{c}_t}$ with COMPR and COMPR\&REG, and $\mathbf{c}_t$ with STATEFUL-BASELINE and REG. For each batch, the high-dimensional state-space was reduced from 128 to 20 dimensions using principal component analysis (PCA) prior to conducting the state-space analyses described below. We refer to the dimension-reduced cell state as $\mathbf{c}_t^*$. On average across instances, the 20 principal components explained 92\,\% (logic evaluation), 94\,\% (search), 97\,\% (shift) and 100\,\% (all other tasks) of the variance.


\paragraph{Curvature of State-Space Trajectories}
We quantified the deviation of a state-space trajectory from a straight line. First, we computed the unit difference vector between adjacent cell states:
\begin{align*}
\mathbf{\Delta_t} &= \mathbf{c}_{t+1}^* - \mathbf{c}_t^* \\
\mathbf{\widehat{\Delta_t}} &= \frac{\mathbf{\Delta_t}}{\|\Delta_t\|}
\end{align*}
Next, we calculated the change in this unit difference vector across the sequence:
\begin{align*}
\kappa_t &= \|\mathbf{\widehat{\Delta_{t+1}}} - \mathbf{\widehat{\Delta_t}}\|
\end{align*}
Cell states along a straight line yield the same unit difference vectors, resulting in $\kappa_t$ being 0. Any deviation from a straight line results in $\kappa_t > 0$. Eventually, we averaged $\kappa_t$ across the entire sequence, including both the encoding and decoding phases, as well as all samples in the batch. We refer to this as curvature $K$ and obtained a single curvature value for each trained DNC instance, i.e., trial.

\paragraph{Patterns of State-Space Trajectories}
We used t-distributed stochastic neighbour embedding (t-SNE) \cite{Maaten2008} to visualise the evolution of the controller states $\mathbf{c}_t^*$ during processing in two dimensions. t-SNE was applied separately to the encoding and decoding phases using the Euclidean metric. The t-SNE implementation from the Python scikit-learn toolbox version 1.0.2 was employed (perplexity was set to 50; learning rate set to "auto"; all other parameters left at default settings).

\subsection{Memory Attention Mechanism Analysis}
\label{sec:memory attention mechanism analysis}
The DNC supports two memory attention mechanisms for writing (content-based addressing and dynamic memory allocation) and two for reading (content-based addressing and backward/forward sequential addressing). We analysed whether our proposed generalisation approaches affect the memory attention mechanisms. To this end, we computed the histograms for the \emph{allocation gate} and \emph{read mode vector}, which the DNC obtains from the interface vector $\mathbf{\xi}_t$ (see \cite{Graves2016} for details). The results can be found in the supplementary materials in Section E
.

\section{Results}
\label{results}

\subsection{DNC Performance}
\label{sec:dnc performance}
We evaluated the performance of our proposed generalisation approaches using input sequence lengths from 2 to 100. Note that we trained with input lengths between 5 and 15. To quantify performance, we evaluated the \emph{hit rate} for the search task and \emph{classification accuracy} for all other tasks.

\begin{figure*}[!h]
	\centering
	\includegraphics[width=0.8\textwidth]{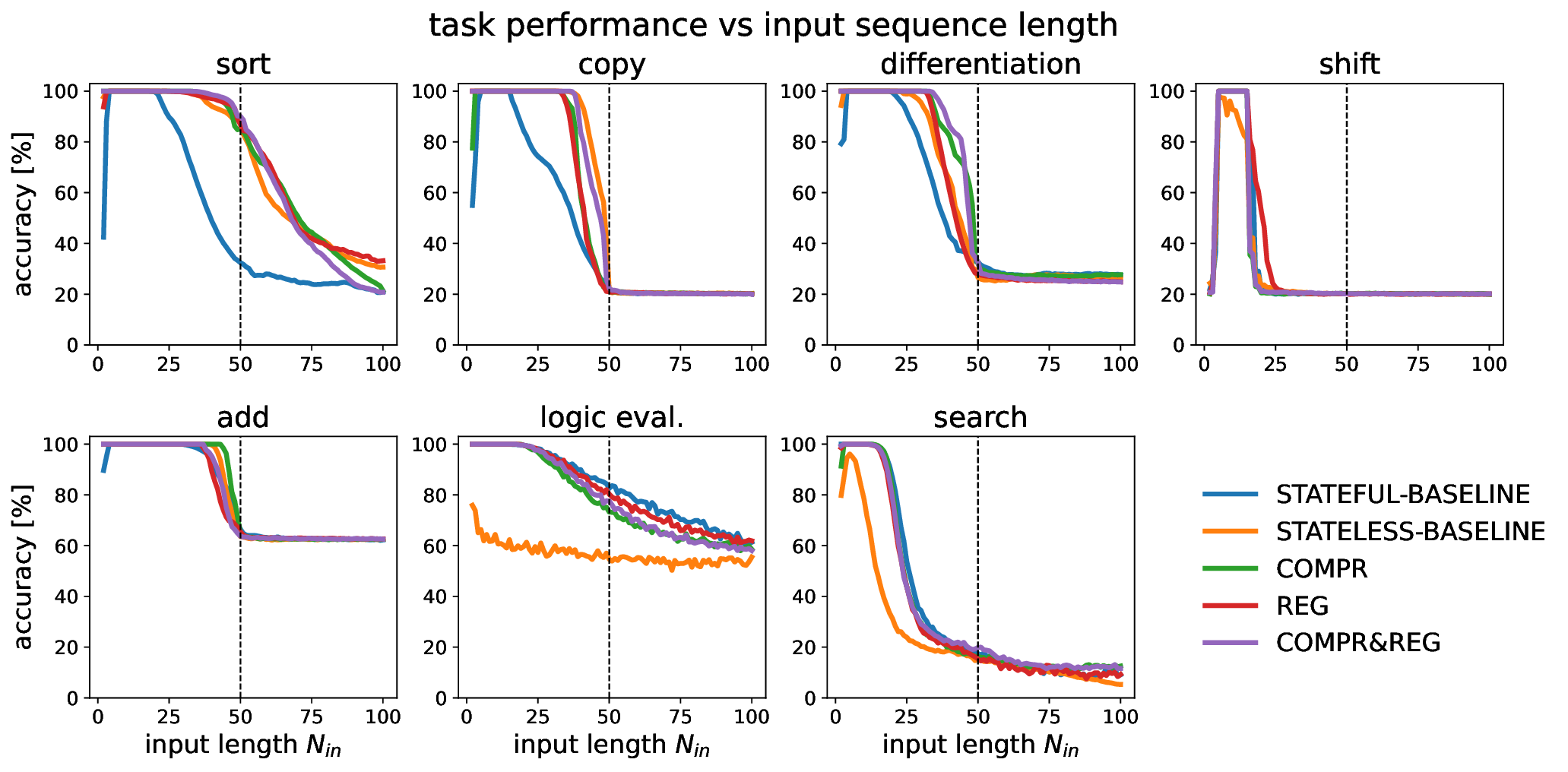}
	\caption{Performances of the generalisation approaches vs the input length for each task. The medians across trials are shown. The dashed vertical line marks the number of memory slots (50).}
	\label{fig:task_performance_vs_input}
\end{figure*}

Fig.~\ref{fig:task_performance_vs_input} shows the performance in relation to the input sequence length for all tasks. Table \ref{table:scores_th_crossing} lists the maximum input sequence lengths that our generalisation approaches and the baselines can process. The maximum processable input length was defined as the longest input sequence length for which the median performance across trials reached at least 95\,\%. The \emph{factor} column was computed by dividing the maximum input length of the best generalisation approach by that of the stateful baseline. Moreover, we calculated for each trial the \emph{average performance} for input sequence lengths from 2 to 45 and provide descriptive and inferential statistics in Table \ref{table:scores_method_stat} and \ref{table:scores_method_test}, respectively. The upper length limit was set below the memory slot size of 50. Inferential statistics were conducted with a two-sided Mann-Whitney U test \cite{Gibbons2010}. To account for multiple comparisons, the false discovery rate (FDR) was controlled at $q=0.05$ \cite{Benjamini1995} and p-values were adjusted \cite{Yekutieli1999}.

\paragraph{Performance with compression and regularisation}
For the \textbf{sort} and \textbf{copy tasks}, our generalisation approaches processed input sequences up to 110\,\% longer compared to the stateful baseline (c.f.\ Table \ref{table:scores_th_crossing}). Additionally, for both tasks, the classification accuracy, averaged from $L_{in}=2$ to 45, was statistically significantly greater for the combined generalisation approach COMPR\&REG than for the stateful baseline. The individual generalisation approaches were furthermore statistically significantly better in the sort task (c.f.\ Table \ref{table:scores_method_test}).

In the \textbf{differentiation task}, our generalisation approaches allowed input sequences to be up to 50\,\% longer and achieved better classification accuracies than the stateful baseline (c.f.\ Tables \ref{table:scores_th_crossing} and \ref{table:scores_method_stat}). However, these classification accuracies were not statistically significantly higher than those of the stateful baseline (c.f.\ Table \ref{table:scores_method_test}).

For the \textbf{shift}, \textbf{add}, \textbf{logic evaluation}, and \textbf{search tasks}, Fig.~\ref{fig:task_performance_vs_input} and Tables \ref{table:scores_th_crossing} and \ref{table:scores_method_stat} demonstrate that our compression and regularisation approaches performed comparably to the stateful baseline. In the logic evaluation and search tasks, the stateless baseline showed a small but statistically significant greater performance than the COMPR and REG approaches, respectively (c.f.\ Tables \ref{table:scores_method_stat} and \ref{table:scores_method_test}). For the shift and search tasks, there was only a minimal generalisation beyond the input sequence lengths encountered during training.

We implemented our generalisation approaches using a Peephole LSTM. For comparison, we also implemented these approaches with a vanilla LSTM and evaluated its performance on the sort task, as detailed in Section C
of the supplementary materials. In fact, the generalisation performance significantly dropped when employing state compression or state regularisation with a vanilla LSTM. Additionally, we compared the Peephole LSTM with the vanilla LSTM without any generalisation approaches and obtained similarly poor generalisation performance. Therefore, while the Peephole LSTM itself did not demonstrate generalisation abilities, it facilitated the generalisation with state compression and regularisation.

\paragraph{Performance of the stateless controller}
Omitting a controller's internal memory improved performance in the \textbf{sort}, \textbf{copy}, \textbf{differentiation}, and \textbf{add tasks} compared to the stateful baseline (see maximum input lengths and average performances in Tables \ref{table:scores_th_crossing} and \ref{table:scores_method_stat}, respectively). For the copy task, the average performance of the stateless baseline was statistically significantly better than that of the stateful baseline (see Table \ref{table:scores_method_test}). In these tasks, the stateless baseline yielded similar performance scores to the generalisation approaches (c.f. Tables \ref{table:scores_th_crossing} and \ref{table:scores_method_stat}) and was not statistically significantly different (c.f.\ Table \ref{table:scores_method_test}).

For the \textbf{shift}, \textbf{logic evaluation}, and \textbf{search tasks}, the stateless baseline performed worse than any other approach. The stateless baseline even failed to process input sequence lengths seen during training (c.f. Table \ref{table:scores_th_crossing}). For the logic evaluation and search tasks, the average performance of the stateless baseline was statistically significantly worse than that of any other approach (c.f. Table \ref{table:scores_method_test}).



\begin{table}[!h]
	\tiny
	\centering
	\caption{Maximum input lengths with a median task performance of at least 95\,\% (hit rate for search, otherwise classification accuracy). The \emph{factor} column shows the improvement ratio, calculated as the maximum input length of the best generalisation approach divided by that of the stateful baseline.}
	\label{table:scores_th_crossing}
	\input{tables/scores_th_crossing.tex}
\end{table}

\begin{table*}[!h]
	\tiny
	\setlength\tabcolsep{2pt}
	\centering
	\caption{Descriptive statistics of the average performance from $L_{in}=2$ to 45 over trials (hit rate for search, otherwise classification accuracy in~\%). The largest median values are bold.}
	\label{table:scores_method_stat}
	\input{tables/scores_method_stat.tex}
\end{table*}

\begin{table*}[!h]
	\tiny
	\centering
	\caption{Statistical comparison of the average performance from $L_{in}=2$ to 45 with a two-sided Mann-Whitney U test. $p$-values were FDR-adjusted. Statistically significant differences are bold ($p \leq 0.05$).}
	\label{table:scores_method_test}
	\input{tables/scores_method_test.tex}
\end{table*}

\subsection{DNC Performance with Extended Memory}
\label{sec:dnc performance ext mem}
\begin{figure*}[!h]
	\centering
	\includegraphics[width=0.8\textwidth]{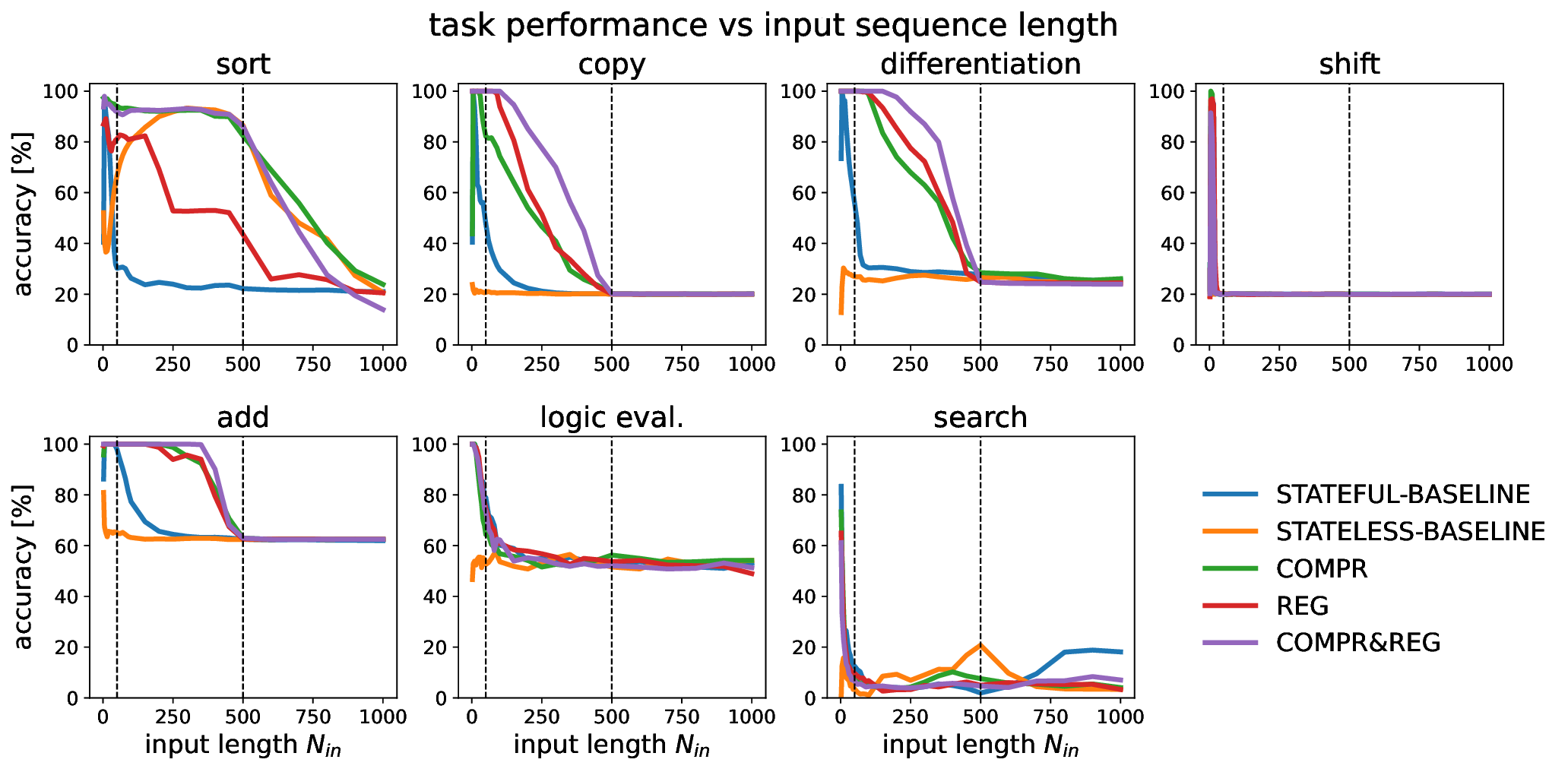}
	\caption{DNC with memory extension: Performances of the generalisation approaches vs the input length for each task. The medians across trials are shown. The dashed vertical lines mark the number of original and extended memory slots, 50 and 500, respectively.}
    
	\label{fig:task_performance_vs_input_extmem}
\end{figure*}

We replaced the memory of a pre-trained DNC with a ten-times-larger memory of size $N = 500$. Performance scores for input sequence lengths ranging from $L_{in} = 2$ to 1000 are shown in Fig.~\ref{fig:task_performance_vs_input_extmem}, and maximal input sequence lengths can be found in Table \ref{table:scores_th_crossing_extmem}. Additional analyses can be found in Section D
in the supplementary materials.

With the extended memory, the length of processable input sequences increased considerably for the \textbf{copy}, \textbf{diff}, and \textbf{add tasks} using the combined generalisation approach (COMPR\&REG), ranging from 780\,\% to 940\,\%.

For the \textbf{sort task}, the generalisation performance was impeded by the memory extension. Nevertheless, the COMPR approach still yielded better generalisation than the stateful baseline of the non-extended DNC. The lengths of the processable input sequences in the \textbf{shift} and \textbf{logic evaluation tasks} were slightly shorter than those of the stateful baseline of the non-extended DNC. In the \textbf{search task}, all approaches failed to process any input sequences.

The stateful baseline's performance decreased in all tasks except the add task due to the memory extension. The stateless baseline failed in all tasks. Therefore, only our generalisation approaches yielded performances similar to or better than those of the stateful baseline of the non-extended DNC. With the COMPR\&REG approach, memory extension was beneficial in the copy, differentiation, and add tasks, and enabled generalisation to considerably longer input sequences.

\begin{table}[!h]
	\tiny
	\centering
	\caption{DNC with memory extension: Maximum input lengths with a median task performance of at least 95\,\% (hit rate for search, otherwise classification accuracy). The best approaches are bold. \emph{factor} was computed between the best generalisation approach and the stateful baseline of the \emph{non-extended DNC}.}
	\label{table:scores_th_crossing_extmem}
	\input{tables/scores_th_crossing_extmem.tex}
\end{table}

\subsection{Curvature of State Trajectories}
\label{sec: curvature results}
\begin{figure*}[!h]
	\centering
	\includegraphics[width=0.6\textwidth]{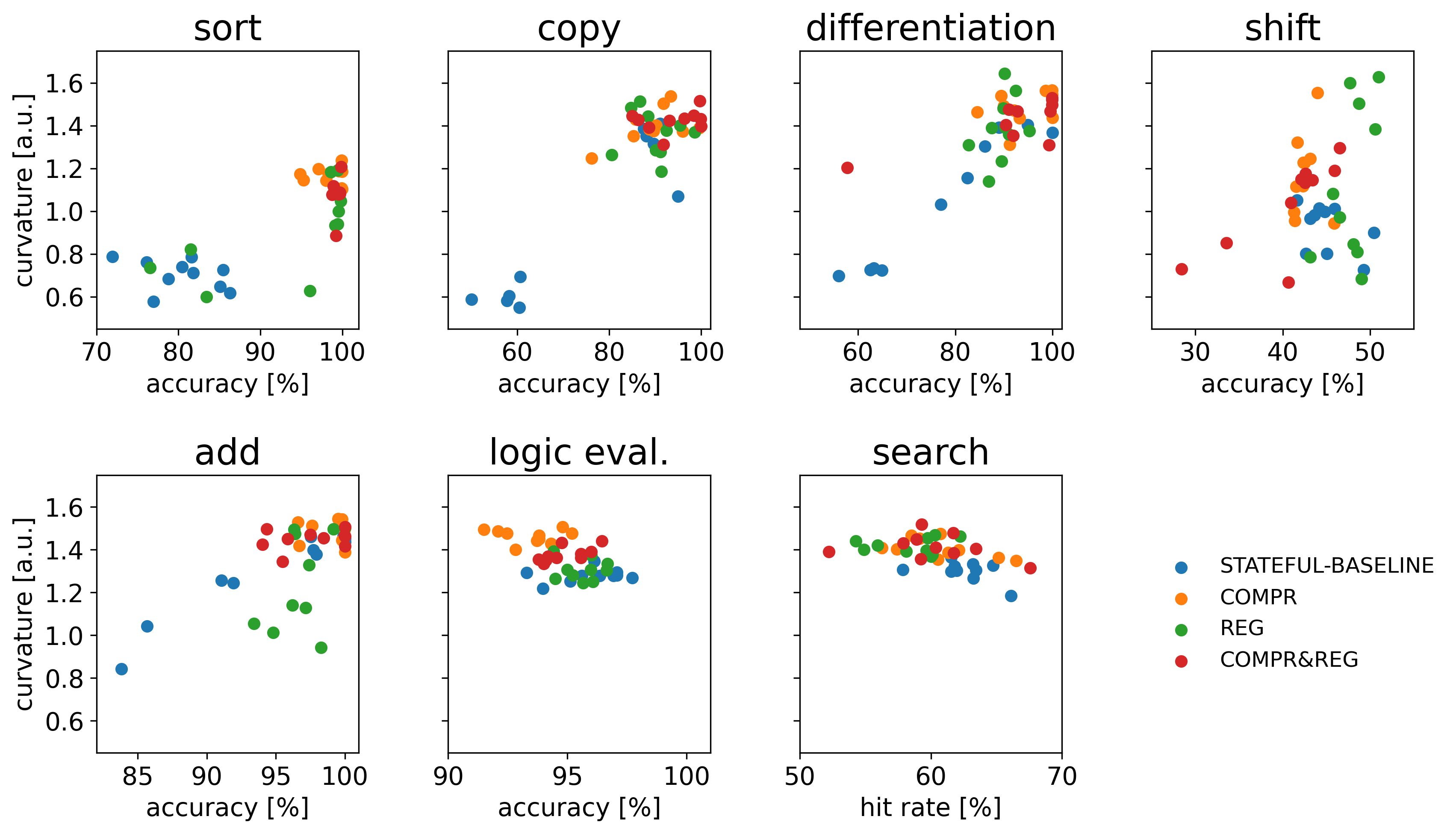}
	\caption{State-space trajectory curvatures and their associated performance scores. Each dot represents one trial (trained DNC instance). Note the different units and scales of the x-axes.}
	\label{fig:controller_states_curvature_vs_score}
\end{figure*}

We presented input sequences of length $L_{in}=30$ to the DNC instances and calculated the curvatures of the state-space trajectories, which reflect the deviation from a straight line. Fig.~\ref{fig:controller_states_curvature_vs_score} shows the relationship between these curvatures and the performance scores presented in \ref{sec:dnc performance} (averaged performance over input sequence lengths of 2 to 45). We then computed the Kendall rank correlation coefficient $\tau$ \cite{Gibbons2010} between the curvatures and performance scores for each task. The Kendall rank correlation coefficient is a robust measure of the ordinal association between two quantities based on rank correlations. We found significant positive correlations for the \textbf{sort}, \textbf{copy}, \textbf{differentiation}, and \textbf{add tasks}, while significant negative correlations were observed for the \textbf{logic evaluation} and \textbf{search tasks} (see Table \ref{table:correlations_curvature_vs_score}; p-values were computed using a two-sided test based on asymptotic normal approximation and adjusted with the Benjamini \& Hochberg procedure \cite{Benjamini1995}). Thus, for sort, copy, differentiation, and add tasks, better performance was linked to greater curvature of the state-space trajectories. Conversely, for the logic evaluation and search tasks, better performance was associated with smaller curvature. However, in these latter tasks, the curvature values were already relatively high and showed minimal variation compared to the other tasks. Overall, a strong increase in curvature was associated with improved performance.

\begin{table}[!h]
	\tiny
	\centering
	\caption{Kendall rank correlation coefficient $\tau$ between state-space trajectory curvatures and performance scores. p-values were adjusted for multiple comparisons \cite{Benjamini1995}. Significant p-values are bold.}
	\label{table:correlations_curvature_vs_score}
	\input{tables/correlations_curvature_vs_score.tex}
\end{table}

\subsection{Patterns of State-Space Trajectories}
\label{sec:controller states}

\begin{figure*}[!h]
	\centering
	\includegraphics[width=0.7\textwidth]{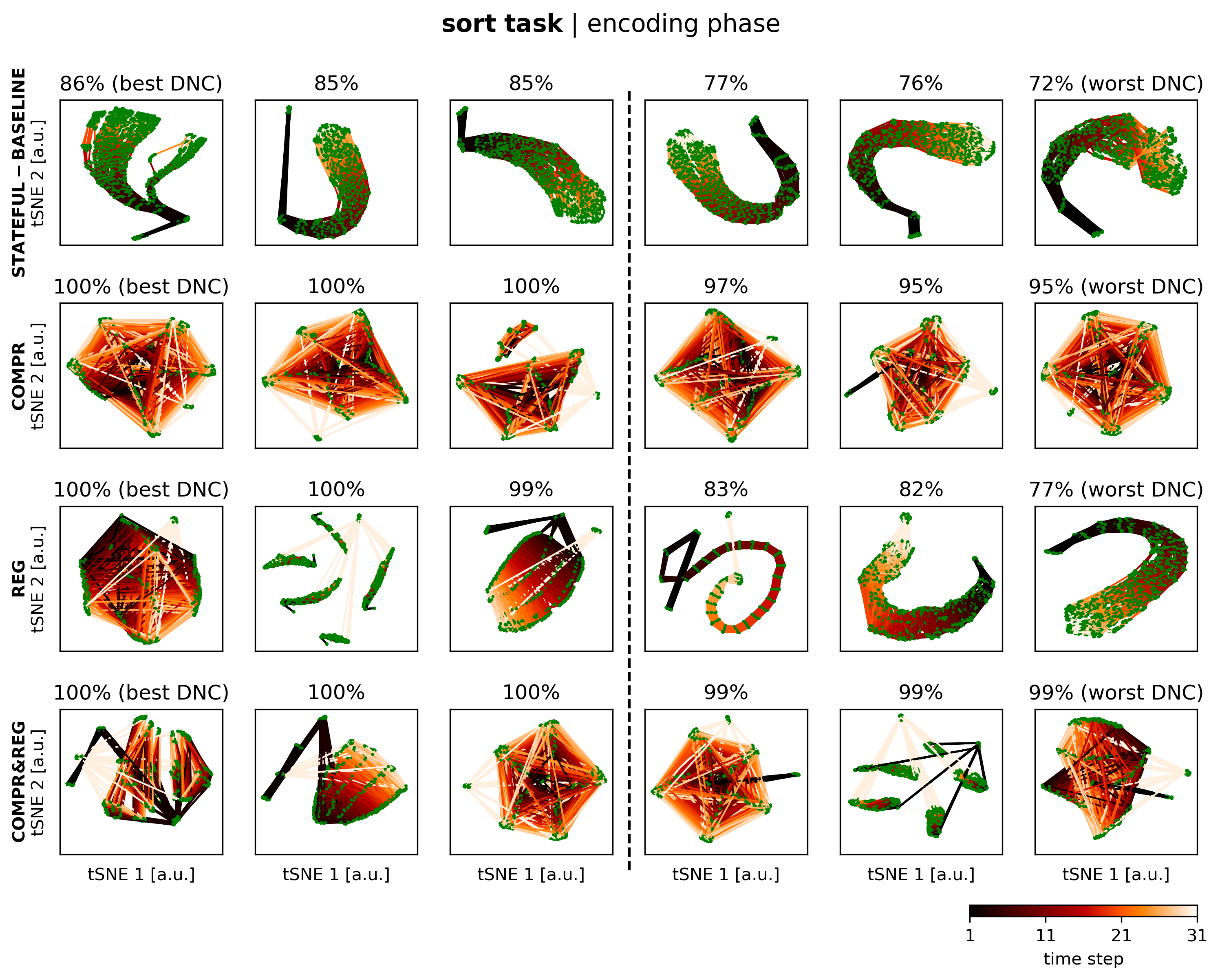}
	\caption{Cell state trajectories from the encoding phase of the sort task projected with t-SNE. Shown are the three best and three worst performing DNC instances (trials) for each generalisation approach (with respect to average performance for input lengths from 2 to 45) on the left and right half, respectively. The trajectories show the processing of a sequence of length 30 and the end-of-input flag. The cell states are marked as green dots with colour-coded transitions (from dark to light in the direction of processing). The trajectories of 64 samples were overlaid for each trial.}
	\label{fig:controller_states_sort_encoding}
\end{figure*}

\begin{figure*}[!h]
	\centering
	\includegraphics[width=0.7\textwidth]{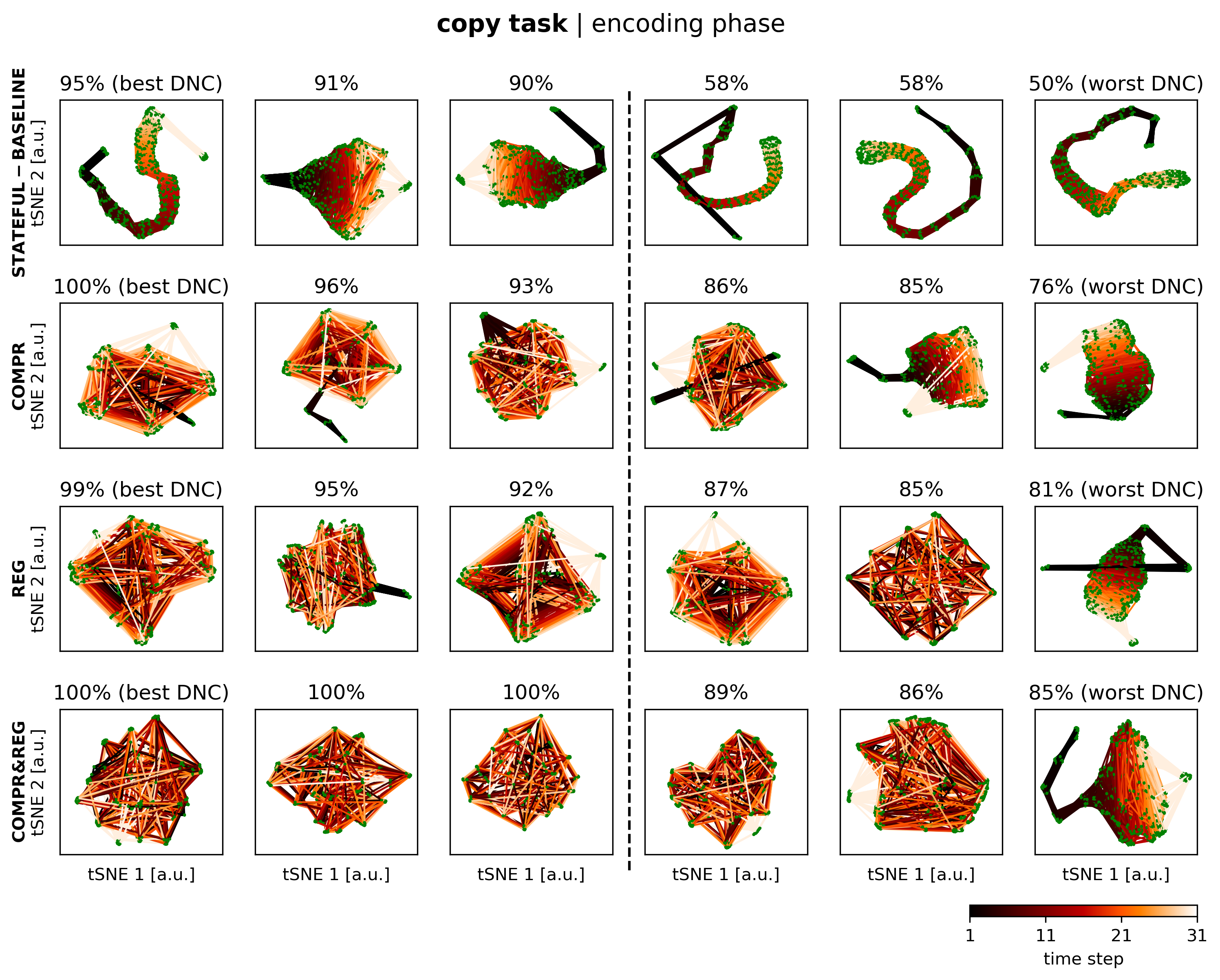}
	\caption{Cell state trajectories from the encoding phase of the copy task projected with t-SNE.}
	\label{fig:controller_states_copy_encoding}
\end{figure*}

Fig.~\ref{fig:controller_states_sort_encoding} and \ref{fig:controller_states_copy_encoding} illustrate the state-space trajectories during the encoding phases of the sort and copy tasks, respectively. The decoding phases and remaining tasks are detailed in the supplementary materials in Section F
.

During the \textbf{encoding phase} of the sort and copy tasks, the stateful baselines often produced state trajectories that followed a largely unidirectional path. In contrast, the generalisation approaches usually yielded clusters in trials with a high performance. Often, five clusters --- the same number as input symbols --- were identifiable, with the states being visited repeatedly over time and forming loops. In the sort task, state regularisation (REG) and the combination approach (COMPR\&REG) also displayed clusters where states evolved in parallel directions over time, and trajectories crossed between clusters (for example, the third-best trial of REG). Additionally, we observed patterns featuring five clusters that were entered based on the initial input numeral, without any crossings between clusters (such as the second-best trial of REG).

State clusters with loops were also observed during the encoding phase of the differentiation, add, and search tasks. Interestingly, while generalisation was possible in the differentiation and add tasks with our approaches (especially with memory extension), generalisation was not seen in the search task. No distinct clusters were identified in the shift and logic evaluation tasks.

In the \textbf{decoding phase}, state clusters were less common. Clusters with loops were only evident in the copy and differentiation tasks when using state regularisation.

Overall, clusters with loops were more pronounced for DNC instances achieving top results. However, the presence of state loops was not a sufficient condition for generalisation.

\section{Discussion}
\label{sec:discussion}
We demonstrated that state-space compression and regularisation can enhance the DNC's ability to generalise to longer input sequence lengths. Interestingly, while the stateless baseline controller failed to learn certain tasks, it facilitated generalisation in others. Given that a stateless controller can be regarded as an extreme form of a state-space constraint, this further reinforces the significance of state-space constraints for generalisation.

\paragraph*{State-Space Structure}
We applied compression and regularisation to create an inductive bias towards state trajectories with greater curvature, potentially leading to the formation of loops. The controller would revisit states rather than drift into areas of state-space that had not been encountered during training. Importantly, our hypothesis aligns with our observations across various tasks, especially during the encoding phases. Trials with top performance often exhibited more structured state spaces, including loops, compared to those with lower performance. Furthermore, stateful baseline trials with top performance also displayed a more structured state space. This indicates that the identified state-space structures were not merely an epiphenomenon of the generalisation approaches but indeed enabled generalisation in most tasks. However, in the search task, state-space loops were not followed by generalisation. Overall, our approaches enhance the emergence of structures like loops, and whether these structures facilitate generalisation depends on the specific tasks.

\paragraph*{Algorithmic Interpretation}
Our generalisation approaches introduce an inductive bias with notable algorithmic consequences. The observed clustered state-space structure resembles a finite state machine (FSM) (c.f.\ \cite{Paassen2022}). In this context, a cell state cluster corresponds to an FSM state, and the network transitions from the current cell state $\mathbf{c}_t$ to the next cell state $\mathbf{c}_{t+1}$ based on the input $\mathbf{\chi}_{t+1}$ (comprising input vector $\mathbf{x_{t+1}}$ and memory readouts $\mathbf{r_{t+1}}$). As the controller network's output $\mathbf{h}_{t+1}$ depends on both the current cell state $\mathbf{c}_t$ and the input $\mathbf{\chi}_{t+1}$, the FSM can be further classified as a Mealy machine \cite{Mealy1955}. However, for this abstraction to hold, the input and output must be clusterable into a finite number of elements.

Our generalisation approaches evidently introduce an inductive bias towards FSM-like behaviour. This could enhance \emph{algorithmic alignment} and, consequently, generalisation for certain tasks \cite{Xu2020,Xu2021}, either by limiting the number of states or by the resulting state graphs (e.g., loops).

\paragraph*{Memory Extension}
We demonstrated that our generalisation methods can enable the use of a larger memory matrix \emph{without retraining}, thereby saving computational resources during training. In four tasks, we were able to process significantly longer input sequences, particularly when combining compression and regularisation. Despite increasing the memory size by a factor of 10, we observed a smaller-than-expected increase in the length of processable input sequences, which varied by task. This might be attributed to the accumulated noise introduced by each memory access operation \cite{Csordas2019}. Unlike our generalisation approaches, both baselines failed to generalise or perform adequately with the extended memory. As the stateless approach represents an extreme form of a state-space constraint, one might expect better generalisation performance, potentially akin to compression or regularisation. However, the absence of a state-space (i.e., an internal memory) means the DNC controller had to depend more on the memory matrix, possibly explaining its sensitivity to variations in memory size.

\paragraph*{Generalisation Approaches}
The original NTM \cite{Graves2014} could generalise to unseen input lengths in certain algorithmic tasks (e.g.\ copy task) without the use of an explicit generalisation method. We also found that the stateful baseline sometimes produced solutions that could generalise. However, our generalisation approach produced solutions that could generalise more frequently. Furthermore, the limitations of the NTM compared to the DNC may have facilitated generalising solutions. For example, Taguchi and Tsuruoka \cite{Taguchi2018} demonstrated that performance could be improved by excluding the controller output when calculating the final output of the DNC or NTM. Rae et al.\ \cite{Rae2016} have shown that sparse memory access can facilitate input-length generalisation in an associative recall task. Hence, model limitations, such as our state-space constraints, can facilitate generalisation.

Model limitations are not the only means of achieving generalisation. Yang et al.\ \cite{Yang2017} potentially increased the complexity of the model by combining absolute addressing with relative addressing based on group actions, thereby obtaining at least 2-fold input-length generalisation. Le et al.\ \cite{Le2020} increased the number of free parameters in the model by enabling the controller to dynamically load weights during the processing of input sequences. This was shown to be particularly beneficial when a task comprises several sequential sub-tasks. Furthermore, the authors reported improved input length generalisation for single algorithmic tasks compared to the NTM, potentially by facilitating the learning of different controller behaviours for the encoding and decoding phases.

Other factors that can enhance generalisation include the initialisation of the memory matrix \cite{Collier2018}, the choice of its architecture (e.g., a graph neural network \cite{Bidokhti2022a}), and inductive biases shaping how content is stored within it \cite{Azarafrooz2022}. Additionally, special learning paradigms, such as multitask learning \cite{Xhonneux2021} or dual reasoning \cite{Numeroso2023}, may facilitate generalisation in a DNC.

Generally, comparing the performance of individual approaches is challenging since the tasks employed or their specific implementations vary. Nevertheless, our compression and regularisation approaches are orthogonal to existing ones and could be used in conjunction. Our approaches are, in principle, applicable to any MANN with a stateful controller.


\paragraph*{Limitations}
First, our work cannot determine whether the improved generalisation results from successful learning of the respective algorithm, or from enhanced robustness in shortcut learning. This is further complicated by noise accumulation \cite{Csordas2019}, which can potentially mask the output of any learned algorithm. Second, our work does not present a general approach for input-length generalisation. Whether generalisation improves depends on the specific task and its underlying algorithm. Third, we focused exclusively on input-length generalisation. Whether our approaches are also advantageous for other types of generalisation was not explored. Finally, due to the calculation of the cosine similarity between every state pair, the time complexity of the state regularisation approach is $\mathcal{O}(t^2)$. For extremely long sequences, it may therefore be necessary to replace the exhaustive pairwise comparison with heuristic-based methods, such as clustering the state space first and then regularizing based on cluster compactness.


\section{Conclusion}
\label{sec:conclusion}
We have demonstrated that state-space constraints, specifically compression and regularisation, enhance the DNC's ability to generalise to longer input sequence lengths. These approaches are not limited to the DNC and can be applied to any MANN with a stateful controller. Notably, our methods allow a DNC to be trained using shorter input sequences, which (1) saves computational resources and (2) enables training when acquiring data is costly. Furthermore, our methods facilitate extending memory after training to improve input-length generalisation when memory is a limiting factor. Interestingly, we have observed that input-length generalisation is often accompanied by state-space structures that align with a finite state machine.

\section*{Acknowledgment}
Know-Center is funded by the Austrian COMET Program --- Competence Centers for Excellent Technologies --- under the auspices of the Austrian Federal Ministry for Innovation, Mobility and Infrastructure, and the Austrian Federal Ministry for Economic Affairs, Energy and Tourism, and by the state of Styria. COMET is managed by the Austrian Research Promotion Agency (FFG). The authors thank Joana Pereira for helpful comments on the manuscript.



\renewcommand*{\bibfont}{\footnotesize}
\printbibliography

\include{supp_materials}

\end{document}

%% file: tables/scores_th_crossing.tex
\begin{tabular}{l|cccccc}
\hline
{} & STATEFUL-BL. & STATELESS-BL. &   COMPR &     REG & COMPR\&REG & factor \\
task            &                   &                    &         &         &           &        \\
\hline
sort            &                23 &                 39 &      45 &      46 &   \bf 48 &    2.1 \\
copy            &                17 &            \bf 41 &      36 &      35 &        39 &    2.3 \\
differentiation &                24 &                 31 &      35 &      34 &   \bf 37 &    1.5 \\
shift           &           \bf 16 &                  8 & \bf 16 & \bf 16 &   \bf 16 &    1.0 \\
add             &                40 &                 43 & \bf 46 &      39 &        41 &    1.2 \\
logic eval.     &           \bf 32 &                  0 &      27 &      31 &        29 &    1.0 \\
search          &           \bf 18 &                  6 & \bf 18 &      17 &        17 &    1.0 \\
\hline
\end{tabular}

%% file: tables/scores_method_stat.tex
\begin{tabular}{l|cccc|cccc|cccc|cccc|cccc}
\hline
{} & \multicolumn{4}{|c}{STATEFUL-BASELINE} & \multicolumn{4}{|c}{STATELESS-BASELINE} & \multicolumn{4}{|c}{COMPR} & \multicolumn{4}{|c}{REG} & \multicolumn{4}{|c}{COMPR\&REG} \\
{} &            median & max &    mean & SD &             median & max & mean & SD &   median & max &    mean & SD &  median & max &    mean & SD &    median & max &    mean & SD \\
task                  &                   &     &         &    &                    &     &      &    &          &     &         &    &         &     &         &    &           &     &         &    \\
\hline
sort (acc)            &                81 &  86 &      80 &  4 &                 98 & 100 &   98 &  2 &       99 & 100 &      98 &  2 &      99 & 100 &      93 &  9 &   \bf 99 & 100 & \it 99 &  0 \\
copy (acc)            &                74 &  95 &      74 & 17 &            \bf 97 & 100 &   88 & 21 &       90 & 100 &      90 &  6 &      91 &  99 &      90 &  5 &        95 & 100 & \it 94 &  6 \\
differentiation (acc) &                80 & 100 &      78 & 14 &                 88 & 100 &   84 & 19 &       93 & 100 & \it 94 &  5 &      90 &  95 &      90 &  3 &   \bf 96 & 100 &      92 & 12 \\
shift (acc)           &                45 &  50 &      45 &  3 &                 40 &  50 &   37 & 11 &       42 &  46 &      43 &  1 & \bf 48 &  51 & \it 48 &  2 &        42 &  47 &      41 &  5 \\
add (acc)             &                98 & 100 &      95 &  6 &                 99 & 100 &   94 & 11 & \bf 100 & 100 & \it 99 &  1 &      97 & 100 &      97 &  2 &        98 & 100 &      98 &  2 \\
logic eval. (acc)     &           \bf 96 &  98 & \it 96 &  1 &                 60 &  87 &   62 & 14 &       94 &  95 &      93 &  1 &      96 &  97 &      96 &  1 &        95 &  96 &      95 &  1 \\
search (hit rate)     &           \bf 63 &  66 & \it 63 &  2 &                 41 &  50 &   41 &  5 &       61 &  64 &      60 &  2 &      60 &  61 &      58 &  2 &        60 &  68 &      60 &  4 \\
\hline
\end{tabular}

%% file: tables/scores_method_test.tex
\begin{tabular}{l|cccc|ccc|cc|c}
\hline
{} & \multicolumn{4}{c|}{STATEFUL-BASELINE vs} & \multicolumn{3}{c|}{STATELESS-BASELINE vs} & \multicolumn{2}{c|}{COMPR vs} &     REG vs \\
{} &   STATELESS-BASELINE &      COMPR &        REG &  COMPR\&REG &                 COMPR &        REG &  COMPR\&REG &        REG & COMPR\&REG &  COMPR\&REG \\
task            &                      &            &            &            &                       &            &            &            &           &            \\
\hline
sort            &           \bf 0.001 & \bf 0.001 & \bf 0.044 & \bf 0.001 &                 0.382 &      0.970 &      0.053 &      0.310 &     0.915 &      0.202 \\
copy            &                0.069 &      0.136 &      0.102 & \bf 0.038 &                 0.283 &      0.283 &      0.822 &      0.970 &     0.264 &      0.202 \\
differentiation &                0.382 &      0.053 &      0.102 &      0.094 &                 0.187 &      0.804 &      0.202 &      0.151 &     0.965 &      0.053 \\
shift           &                0.283 &      0.052 &      0.102 &      0.094 &                 0.842 &      0.069 &      0.915 & \bf 0.003 &     0.879 & \bf 0.004 \\
add             &                0.879 &      0.296 &      0.970 &      0.428 &                 0.414 &      0.602 &      0.751 &      0.068 &     0.618 &      0.668 \\
logic eval.     &           \bf 0.001 & \bf 0.012 &      0.447 &      0.173 &            \bf 0.001 & \bf 0.001 & \bf 0.001 & \bf 0.004 &     0.053 &      0.151 \\
search          &           \bf 0.001 &      0.136 & \bf 0.006 &      0.187 &            \bf 0.001 & \bf 0.001 & \bf 0.001 &      0.187 &     0.970 &      0.414 \\
\hline
\end{tabular}

%% file: tables/scores_th_crossing_extmem.tex
\begin{tabular}{l|cccccc}
\hline
{} & STATEFUL-BL. & STATELESS-BL. &   COMPR &     REG & COMPR\&REG & factor \\
task            &                   &                    &         &         &           &        \\
\hline
sort            &                10 &                  0 & \bf 40 &       0 &        15 &    1.7 \\
copy            &                10 &                  0 &      35 &     100 &  \bf 150 &   8.8 \\
differentiation &                20 &                  0 &     150 &     150 &  \bf 250 &   10.4 \\
shift           &           \bf 15 &                  0 & \bf 15 & \bf 15 &         0 &    0.9 \\
add             &                60 &                  0 &     350 &     250 &  \bf 400 &    10.0 \\
logic eval.     &           \bf 25 &                  0 &      20 & \bf 25 &        20 &    0.8 \\
search          &                 0 &                  0 &       0 &       0 &         0 &    - \\
\hline
\end{tabular}

%% file: tables/correlations_curvature_vs_score.tex
\begin{tabular}{l|cc}
\hline
{} &   $\tau$ &    $p$ \\
task            &       &            \\
\hline
sort            &  0.41 & \bf 0.001 \\
copy            &  0.28 & \bf 0.014 \\
differentiation &  0.46 & \bf 0.000 \\
shift           &  0.09 &      0.428 \\
add             &  0.32 & \bf 0.006 \\
logic eval.     & -0.35 & \bf 0.003 \\
search          & -0.32 & \bf 0.006 \\
\hline
\end{tabular}

%% file: supp_materials.tex
\clearpage
\setcounter{page}{1}

\section*{Supplementary Materials}

\subsection{Task Implementation Details}
\label{sec:task implementation details}

\paragraph{sort, copy, differentiation, and shift tasks}
For these tasks, the input to the DNC was a two-dimensional vector sequence $\mathbf{x}_t \in \mathbb{R}^2$. The first dimension of $\mathbf{x}_t$ corresponded to the task input sequence $in_t$, the second dimension was a control channel $ctr_t$ indicating the end of the input. The DNC output was also a two-dimensional vector $\mathbf{y}_t \in \mathbb{R}^2$, where the first dimension corresponded to the task result $out_t$ and the second dimension was a signal channel $sig_t$ indicating the end of the DNC output. The DNC first received the task input $in_t$ for $L_{in}$ time steps (with $in_t = 0, \forall t > L_{in}$). After the task input was presented to the DNC (i.e.,\ the input was encoded), the control channel $ctr_t$ was switched from 0.0 to 1.0 at $t = L_{in} + 1$ only. This was the \emph{end-of-input flag} which marked both the end of the task input and the encoding phase, and indicated the beginning of the decoding phase. The task result $out_t$ was then decoded by the DNC for $L_{out}$ time steps, i.e.\ from $t = L_{in} + 2$ to $t = L_{in} + 1 + L_{out}$. The DNC was trained to set the signal channel $sig_t$ to 1.0 at $t = L_{in} + 1 + L_{out} + 1$, and to 0.0 before this time (\emph{end-of-output flag}). The task result had the same length as the task input, i.e.\ $L_{out} = L_{in}$ (for the differentiation task, we added a constant symbol to the task result). Thus, the total number of time steps to process these tasks was $N_{total} = L_{in} + 1 + L_{out} + 1 = 2L_{in} + 2$. 

We evaluated these tasks with a base-5 numeral system, i.e.\ a quinary system, comprising numerals 0 to 4. These numerals were not used directly as task input $in_t$ but were normalised to the interval $[-1.0, 1.0]$. In the training phase, we also normalised the target for the task result $out_t$ to the same interval. In the evaluation phase, $out_t$ was de-normalized and rounded.

\paragraph{add task}
The add task was implemented using three-dimensional input and output vector sequences $\mathbf{x}_t \in \mathbb{R}^3$ and $\mathbf{y}_t \in \mathbb{R}^3$. Analogous to the tasks described previously, the last dimension of $\mathbf{x}_t$ and $\mathbf{y}_t$ corresponded to the control channel $ctr_t$ and the signal channel $sig_t$, respectively. The first two dimensions of $\mathbf{x}_t$ encompassed the two task input sequences $in_{1,t}$ and $in_{2,t}$, which had to be added by the DNC. It was a binary add task, and the binary numerals 0 and 1 were encoded as -1.0 and 1.0, respectively, in $in_{1,t}$ and $in_{2,t}$. The first two dimensions of $\mathbf{y}_t$ encompassed the binary addition result ranging from 00 to 10 (the output numerals 0 and 1 were encoded as -1.0 and 1.0, respectively). The control channel $ctr_t$ and the signal channel $sig_t$ were used as in the tasks described above and marked the end of the encoding and decoding phases. The total number of time steps for the add task was $N_{total} = 2L_{in} + 2$.

\paragraph{search task}
The search task consisted of three phases: encoding, search query, and decoding. In the encoding phase, an input sequence $in_{t}$, consisting of five numerals (0 to 4), was presented to the DNC. Subsequently, the search query $q$ corresponding to one of the five numerals was presented. Eventually, the DNC was supposed to return the positions $out_t$ of the queried numerals within the input sequence.

The total input to the DNC was a two-dimensional vector sequence $\mathbf{x}_t \in \mathbb{R}^2$ where the first dimension contained the input sequence $in_{t}$ with a length of $L_{in}$ followed by a blank and the search query $q$ at $t = L_{in} + 2$. The numerals were normalised to the interval $[0.0, 1.0]$. The second dimension contained the control channel $ctr_t$, which marked the end of the input sequence $in_{t}$, as well as the end of the search query $q$ with a value of 1.0 at both time steps and was set to 0.0 otherwise. The DNC output was a two-dimensional vector sequence $\mathbf{y}_t \in \mathbb{R}^2$, comprising the search results $out_t$ and the signal channel $sig_t$, which marked the end of the search results. $out_t$ yielded the positions that matched the query $q$, with one position per time step and starting from $t = L_{in} + 4$. The positions were normalised to the interval $[0.0, 1.0]$, corresponding to the beginning and end of $in_{t}$, respectively. With $L_{out}$ being the number of found positions, the total number of time steps for the search task was $N_{total} = L_{in} + 3 + L_{out} + 1$.

\paragraph{logic evaluation task}
The DNC was trained to parse a propositional logic formula and infer its truth value. For implementation, we did not use proposition symbols, e.g.\, the $A$ and $B$ in $\neg (A \land B)$, but used true/false values ($\top, \bot$) instead. The problem of defining propositions and assigning logical values to them was outside the scope of this task, and we focused on the actual evaluation of a logic formula. A logical formula in our logic task was, for example, $((\neg (\top \land \bot) \to \top) \lor \bot) \leftrightarrow \top$, which evaluates to $\top$ (true).

The input $\mathbf{x}_t \in \mathbb{R}^{10}$ to the DNC comprised a vector $in_{t} \in \mathbb{R}^9$ representing the logic formula of length $L_{in}$ and a control channel $ctr_t$ separating the encoding and decoding phases. We used one-hot encoding for $in_{t}$ to represent logical values ($\top, \bot$), operators ($\neg, \lor, \land, \to, \leftrightarrow$) and parentheses. The truth value of the logic formula was encoded in the DNC output $\mathbf{y}_t \in \mathbb{R}$ at $t = L_{in} + 2$ as 0.0 or 1.0. The total number of time-steps for this task was $N_{total} = L_{in} + 2$.

\subsection{Neural Network Training}
\label{sec:neural network training}
All tasks, except logic evaluation, were set up as regression problems, where we used the mean squared error (MSE) as the loss function. We averaged the MSE across all dimensions of the DNC output $\mathbf{y}_t$ and the relevant time steps of the decoding phase containing the task result, i.e.\ $N_{total} - L_{out} \leq t \leq N_{total}$. The logic evaluation task was set up as a classification problem, where we used cross-entropy as the loss function for $\mathbf{y}_{N_{total}}$. For each task, we programmatically generated task-specific training samples $(\mathbf{x}_t, \mathbf{y}_t)$. The input length $L_{in}$ of the generated training samples ranged from 5 to 15 (uniformly distributed between batches, but equal for all samples within a batch).

We trained the LSTM, Peephole LSTM, and FFNN-based controllers separately for each task using ADAM \cite{Kingma2015}. We set the learning rate and other ADAM-specific parameters to the values suggested in \cite{Kingma2015}. In particular, the learning rate was set to $\eta = 0.001$. The training encompassed the controller network parameters $\theta$, the output/readout/interface matrices $\mathbf{W}_{\{y,r,\xi\}}$ with $\mathbf{b}_{\{y,\xi\}}$, and, if applicable, the initial states $\mathbf{h}_{0}$ and $\mathbf{c}_{0}$ of the controller network. 
We initialised the parameters $\theta$ of the LSTM-based controllers and the output/readout/interface matrices with the LeCun method \cite{LeCun1998}. For the FFNN, we used the Glorot method \cite{Glorot2010}. We drew the values of the initial cell and hidden state vectors from a centred normal distribution with unit standard deviation.
To prevent the exploding gradient problem, we applied gradient clipping \cite{Pascanu2013}.

Using a batch size of 64, we trained the DNC controllers for 300\,000 iterations. Every \nth{10} iteration, we computed the loss of the DNC using an input length $L_{in} = 30$ to estimate the generalisation or out-of-distribution performance.
We then calculated a running average over the last 500 out-of-distribution losses. Eventually, we selected the iteration with the lowest average out-of-distribution loss for further evaluation.

\subsection{LSTM vs Peephole LSTM}
\label{sec:lstm vs peehole lstm}
\begin{figure}[!h]
	\centering
	\includegraphics[width=0.5\textwidth]{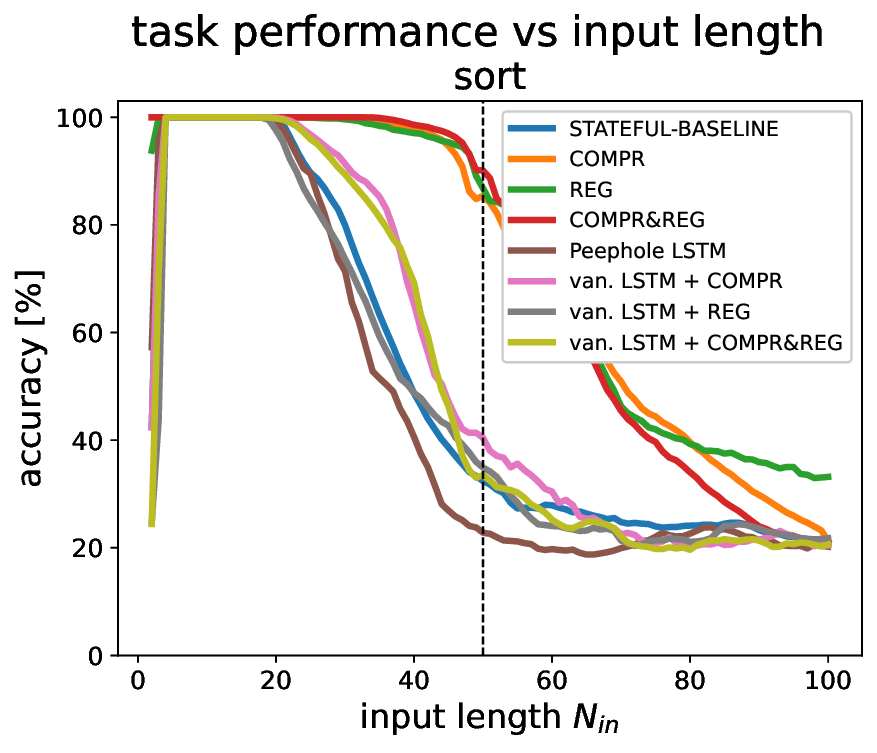}
	\caption{Performance on the sort task when combining state-space constraints with a \emph{Peephole LSTM} (COMPR, REG, COMPR\&REG, as in the main paper) or a \emph{vanilla LSTM} (van.~LSTM + approach). In addition, the performances of the Peephole LSTM and the vanilla LSTM without any state-space constraints are shown (STATEFUL-BASELINE and Peephole LSTM).}
	\label{fig:lstm vs peephole lstm}
\end{figure}

We evaluated the sort task performance when combining compression and regularisation with a \emph{vanilla LSTM} instead of a \emph{Peephole LSTM}, and show the results in Fig.~\ref{fig:lstm vs peephole lstm}. The plot indicates that the architectural changes in the Peephole LSTM are indeed relevant for our generalisation approaches (at least in the sort task), as only the combination of the Peephole LSTM with the state-space constraints considerably improved generalisation performance.

\subsection{DNC Performance with Extended Memory}
\label{sec:dnc performance with extended memory}
Table \ref{table:scores_method_stat_extmem} shows descriptive statistics of the trial performances averaged between $L_{in} = 2$ to $450$. In addition, Mann-Whitney U tests were conducted on the averaged trial performances. The FDR was controlled at $q=0.05$ \cite{Benjamini1995}, adjusted p-values are reported in Table \ref{table:scores_method_test_extmem}.

\begin{table*}[!h]
	\tiny
	\centering
	\caption{DNC with memory extension and descriptive statistics of the task performance over trials (hit rate for search, otherwise classification accuracy, both in \%). The largest median values are bold.}
	\label{table:scores_method_stat_extmem}
	\input{tables/scores_method_stat_extmem.tex}
\end{table*}

\begin{table*}[!h]
	\tiny
	\centering
	\caption{DNC with memory extension: Generalisation approaches compared with a Mann-Whitney U test and FDR-adjusted $p$-values. Statistically significant differences are bold ($p \leq 0.05$).}
	\label{table:scores_method_test_extmem}
	\input{tables/scores_method_test_extmem.tex}
\end{table*}

\subsection{Memory Attention Mechanisms}
\label{sec:memory access modes}
\begin{figure*}[!h]
	\centering
	\includegraphics[width=0.6\textwidth]{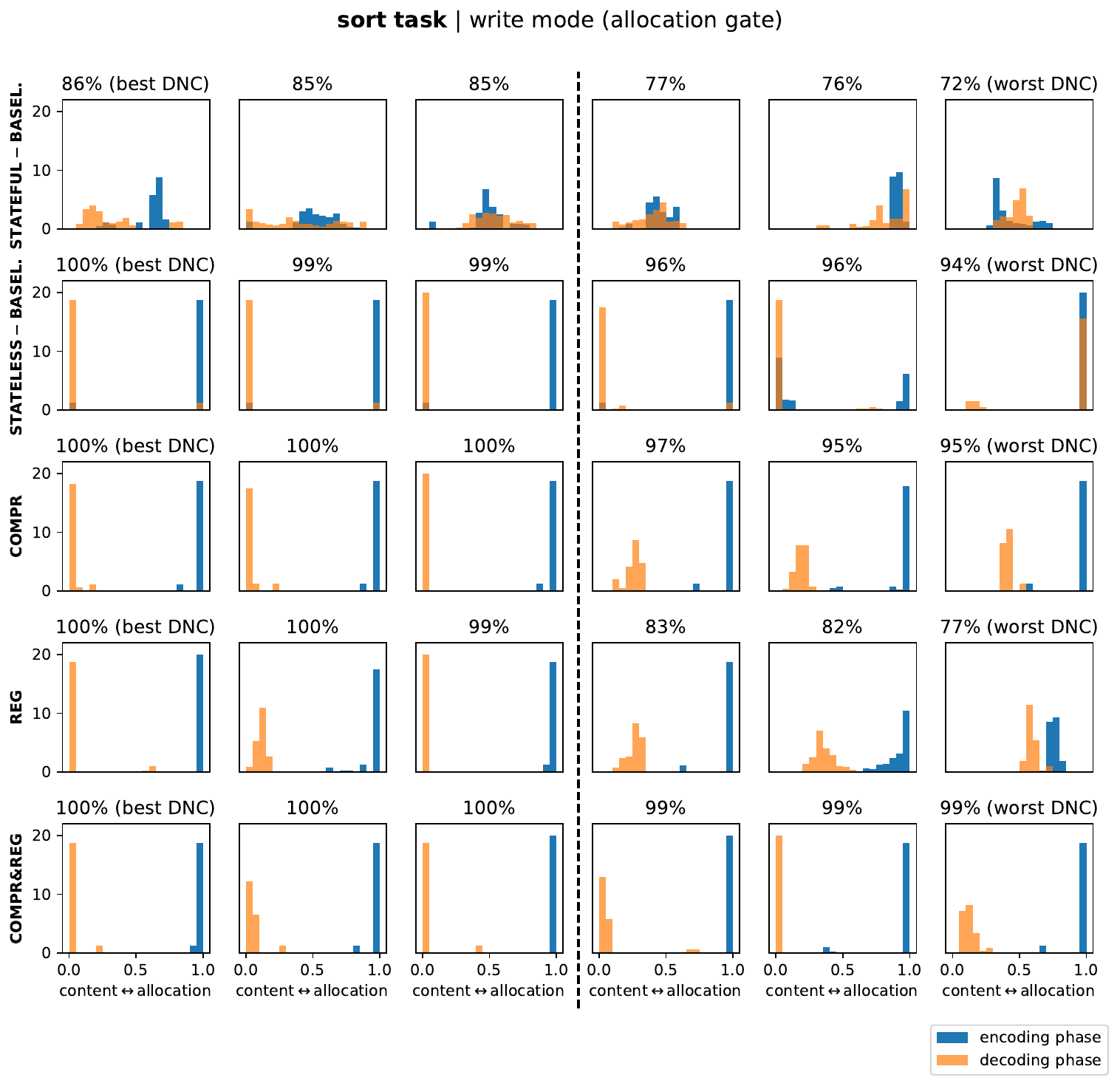}
	\caption{Sort task write modes: Histograms of allocation gate values for input sequence length $L_{in} = 15$. Left/right half: trials with best/worst performance. Values close to 0 indicate content-based addressing, values close to 1 indicate dynamic memory allocation. The encoding and decoding phases are shown separately.}
	\label{fig:memory_access_sort_write}
\end{figure*}

\begin{figure*}[!h]
	\centering
	\includegraphics[width=0.6\textwidth]{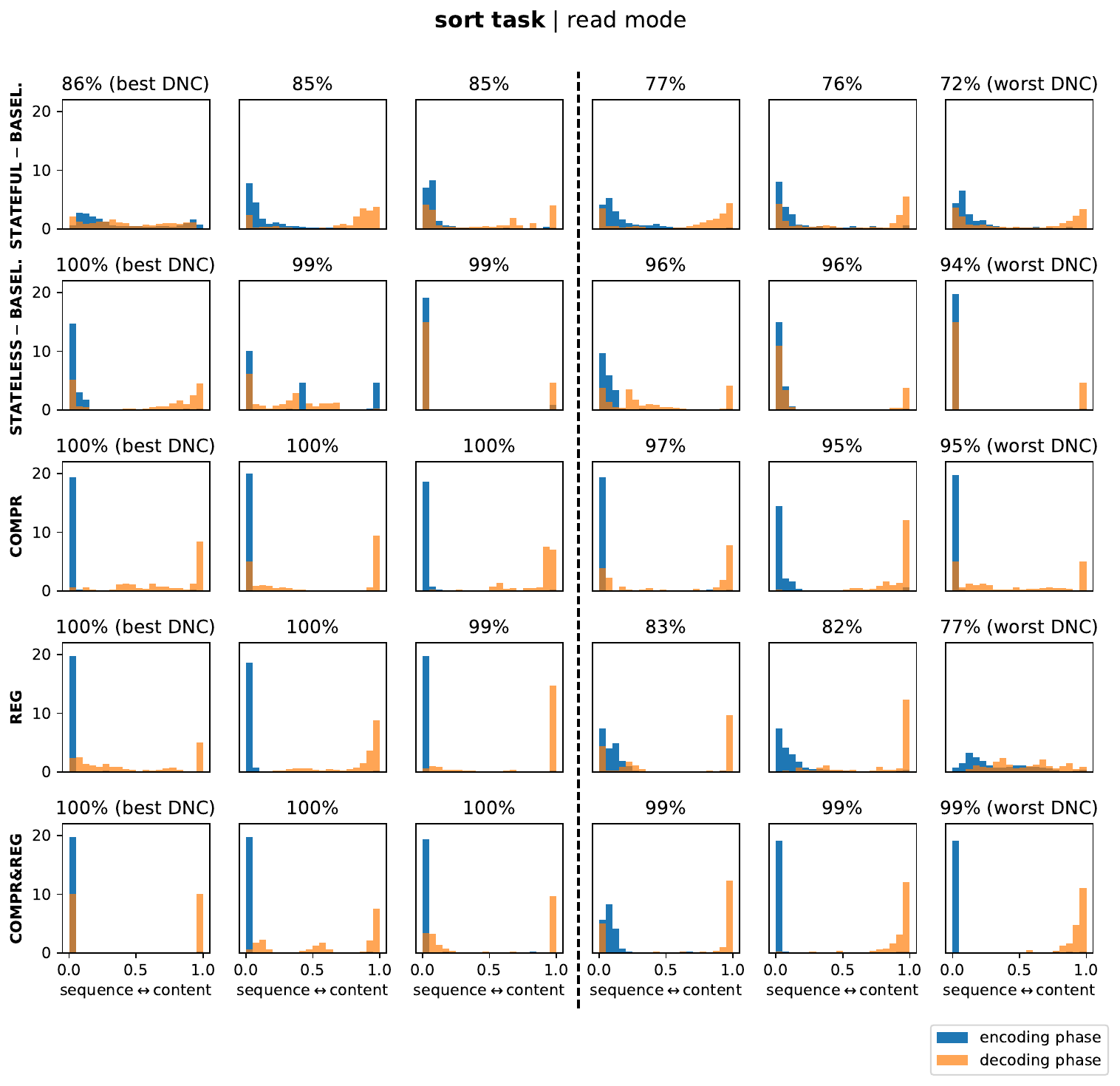}
		\caption{Sort task read modes: Histograms of read mode values for input sequence length $L_{in} = 15$. Left/right half: trials with best/worst performance. Values close to 0 indicate sequential addressing (forward or backward), values close to 1 indicate content-based addressing. The encoding and decoding phases are shown separately.}
	
	\label{fig:memory_access_sort_read}
\end{figure*}

\begin{figure*}[!h]
	\centering
	\includegraphics[width=0.6\textwidth]{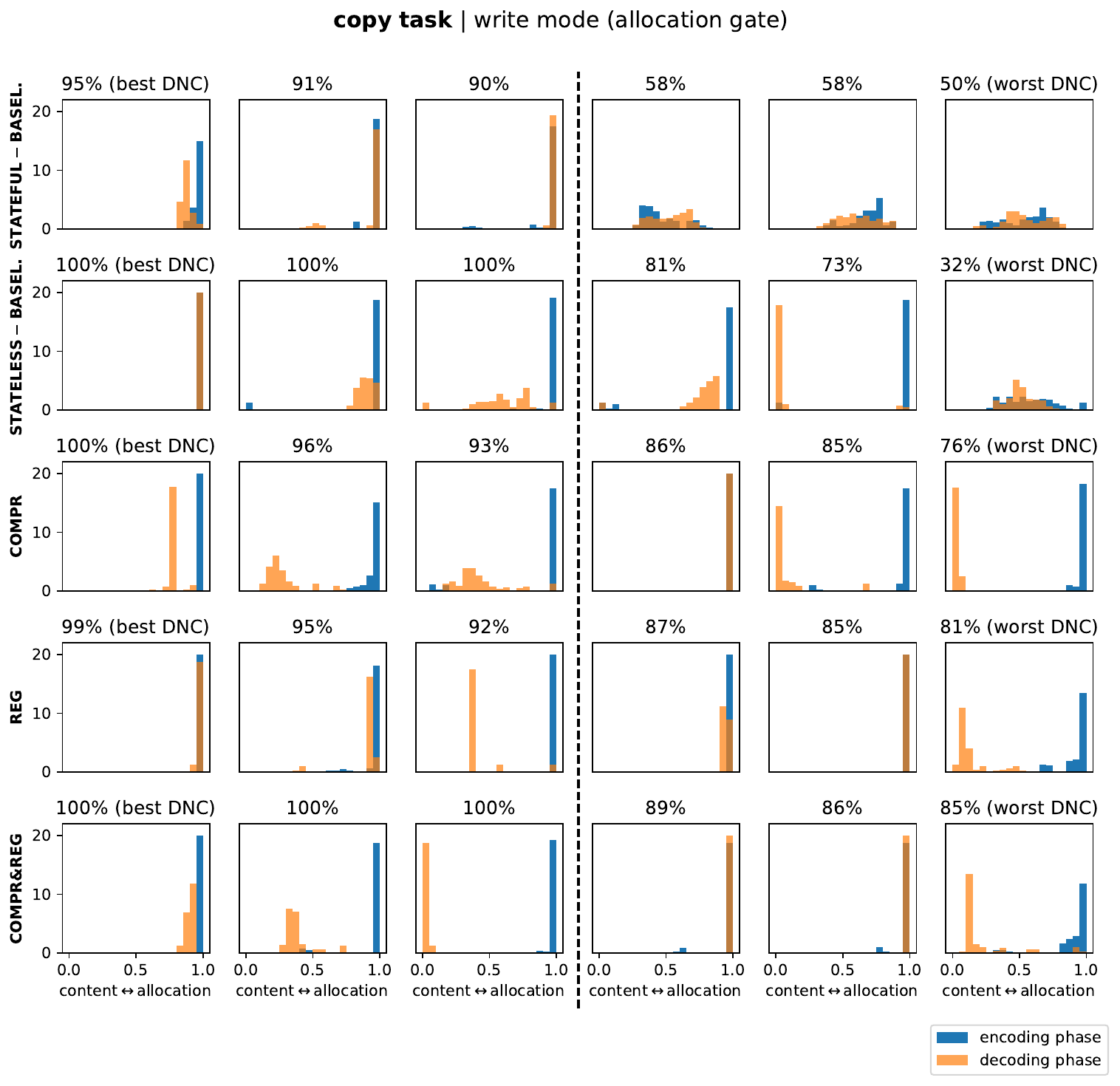}
	\caption{Copy task write modes: Histograms of allocation gate values for input sequence length $L_{in} = 15$. Left/right half: trials with best/worst performance. Values close to 0 indicate content-based addressing, values close to 1 indicate dynamic memory allocation. The encoding and decoding phases are shown separately.}
	\label{fig:memory_access_copy_write}
\end{figure*}

\begin{figure*}[!h]
	\centering
	\includegraphics[width=0.6\textwidth]{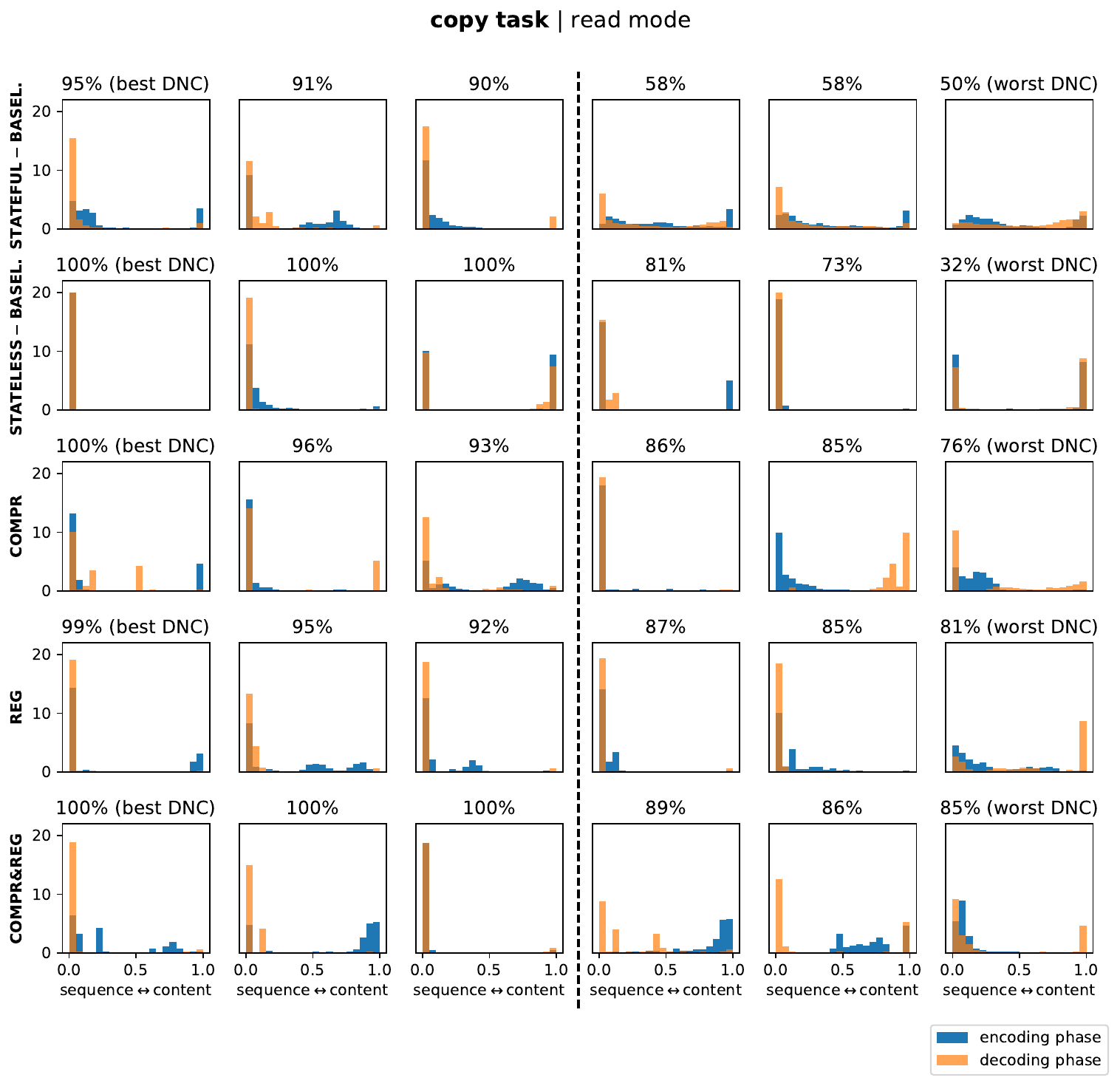}
	\caption{Copy task read modes: Histograms of read mode values for input sequence length $L_{in} = 15$. Left/right half: trials with best/worst performance. Values close to 0 indicate sequential addressing (forward or backward), values close to 1 indicate content-based addressing. The encoding and decoding phases are shown separately.}
	\label{fig:memory_access_copy_read}
\end{figure*}

The DNC supports two memory attention mechanisms for writing (content-based addressing and dynamic memory allocation) and two for reading (content-based addressing and backward/forward sequential addressing). It learns to use and combine these mechanisms in order to solve a task. We analysed whether our generalisation approaches affected the memory attention strategies learnt.
We calculated the histograms of the \emph{allocation gate} and \emph{read mode} vectors. Both vectors are computed at each time step from the interface vector $\mathbf{\xi}_t$, see \cite{Graves2016} for details. The allocation gate sets the ratio between content-based addressing (values close to 0) and dynamic memory allocation (values close to 1) when writing to memory. The read mode vector comprises three components, specifying the ratio between forward sequential addressing, backward sequential addressing, and content-based addressing. We evaluated only the \emph{content-based addressing} component, where values close to 1 corresponded to content-based addressing, and values close to 0 corresponded to sequential addressing in any direction when reading from memory. We refer to the allocation gate value as \emph{write mode}, and to the value of the content-based addressing component of the read mode vector as \emph{read mode} (i.e.\ $g_t^a$ and $\pi_t^i[2]$, respectively, in \cite{Graves2016}). We analysed exemplarily the sort and copy tasks, and show the histograms of the read/write modes for STATEFUL-BASELINE, STATELESS-BASELINE, COMPR, REG, and COMPR\&REG in Figures \ref{fig:memory_access_sort_write} -- \ref{fig:memory_access_copy_read}. The figures show the histograms arranged according to the three best and three worst performing DNC instances (with respect to average performance on input lengths from 2 to 45). The actual histograms were calculated based on an input sequence of length 15, and are shown separately for the encoding and decoding phases. The read modes from the four read heads were aggregated.

With STATEFUL-BASELINE and the sort task, one can observe that the two write modes were often intermixed. Thus, content-based addressing and dynamic memory allocation were used simultaneously when accessing the memory. The encoding and decoding phases showed a similar behaviour. However, the other methods --- which also yielded better generalisation performance --- showed a clear separation between the two write modes. With them, the DNC used predominantly dynamic memory allocation during the encoding phase and content-based writing during the decoding phase. The read mode showed a similar behaviour: the read modes of STATEFUL-BASELINE were intermixed, whereas the read modes of the other methods were well separated. The methods, except STATEFUL-BASELINE, used predominantly sequential reading during the encoding phase and content-based reading during the decoding phase. This could reflect a strategy where the DNC stored the input data via dynamic memory allocation during the encoding phase and performed the actual sorting task in the decoding phase.
Furthermore, DNCs trained with methods other than STATEFUL-BASELINE also tended to intermix read/write modes when the achieved performance was low. Thus, the intermixing of access modes was associated with the STATEFUL-BASELINE method and/or suboptimal performance.

An intermixing of read and write modes can also be observed for the copy task in most DNC instances with low performance. For DNC instances with high performance, the write mode during the encoding phase was predominantly dynamic memory allocation. However, for the decoding phase, the dynamic memory allocation was often mixed with content-based addressing. This was reversed for the read mode. During the encoding phase, content-based addressing was often intermixed with sequential addressing; and during the decoding phase, mostly sequential addressing was used. The predominant use of dynamic memory allocation during the encoding phase and sequential reading during the decoding phase suggests a continuous and steady memory access during both phases, which is expected for a copy task.

To conclude, DNC instances with high generalisation performance often had constant ratios of the respective memory access modes (i.e.\ distinctive peaks in the histograms). Thus, their memory access patterns were rather stable and relatively independent of the actual input data. In contrast, controller networks with low generalisation performance often exhibited ratios that were more spread. Thus, their memory access modes varied throughout the encoding and decoding phases. This could indicate non-robust controller behaviour, highly tuned to the training dataset, which prevents generalisation.

\subsection{Controller States}
\label{sec:controller states sup mat}
\begin{figure*}[!h]
	\centering
	\includegraphics[width=0.7\textwidth]{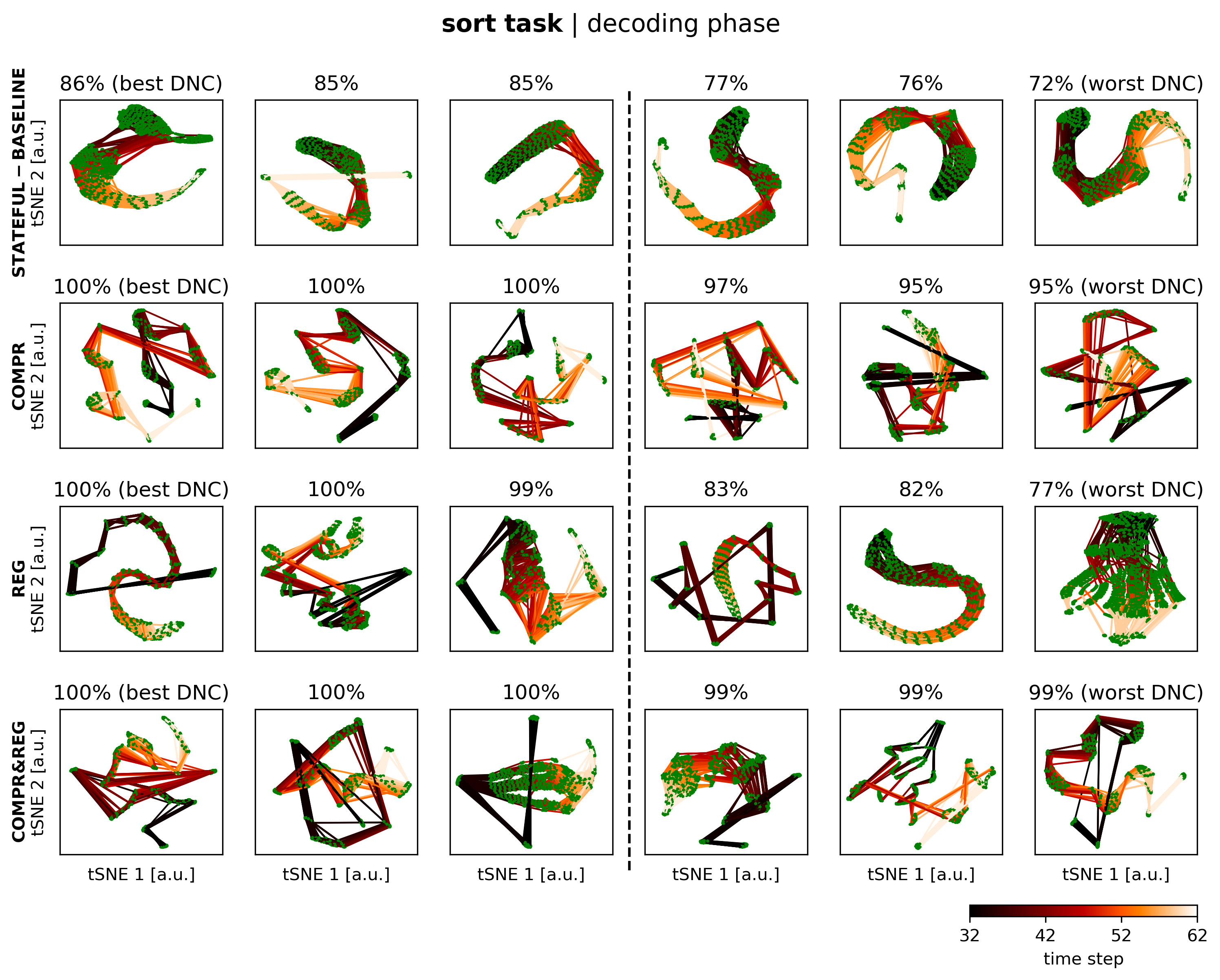}
	\caption{Cell state trajectories from the decoding phase of the sort task projected with t-SNE. Shown are the three best and three worst performing DNC instances (trials) for each generalisation approach (with respect to average performance for input lengths from 2 to 45) on the left and right half, respectively. The trajectories show the processing of a sequence of length 30 and the end-of-output flag. The cell states are marked as green dots with colour-coded transitions (from dark to light in the direction of processing). The trajectories of 64 samples were overlaid for each trial.}
	\label{fig:controller_states_sort_decoding}
\end{figure*}

\begin{figure*}[!h]
	\centering
	\includegraphics[width=0.7\textwidth]{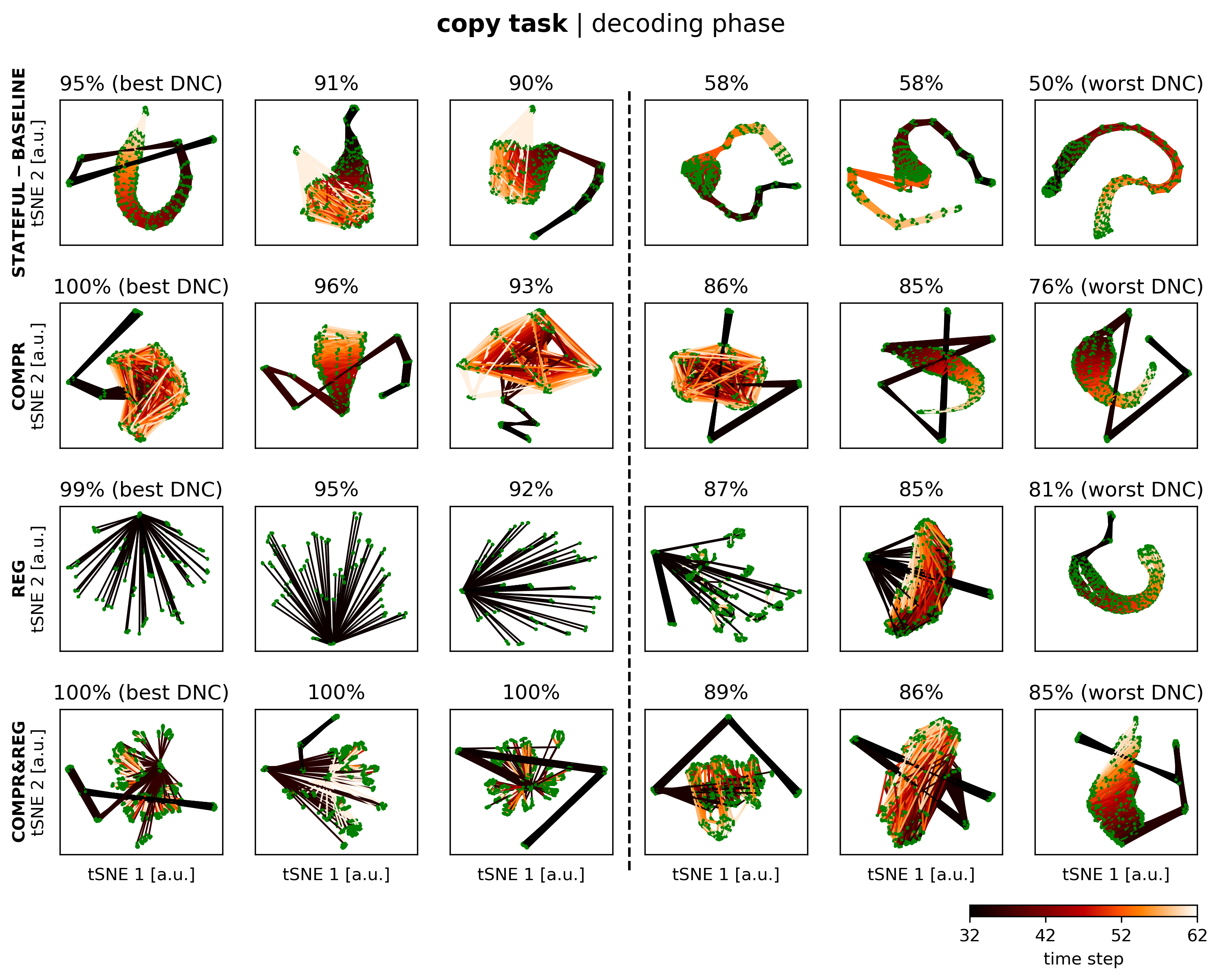}
	\caption{Cell state trajectories from the decoding phase of the copy task projected with t-SNE.}
	\label{fig:controller_states_copy_decoding}
\end{figure*}

\begin{figure*}[!h]
	\centering
	\includegraphics[width=0.7\textwidth]{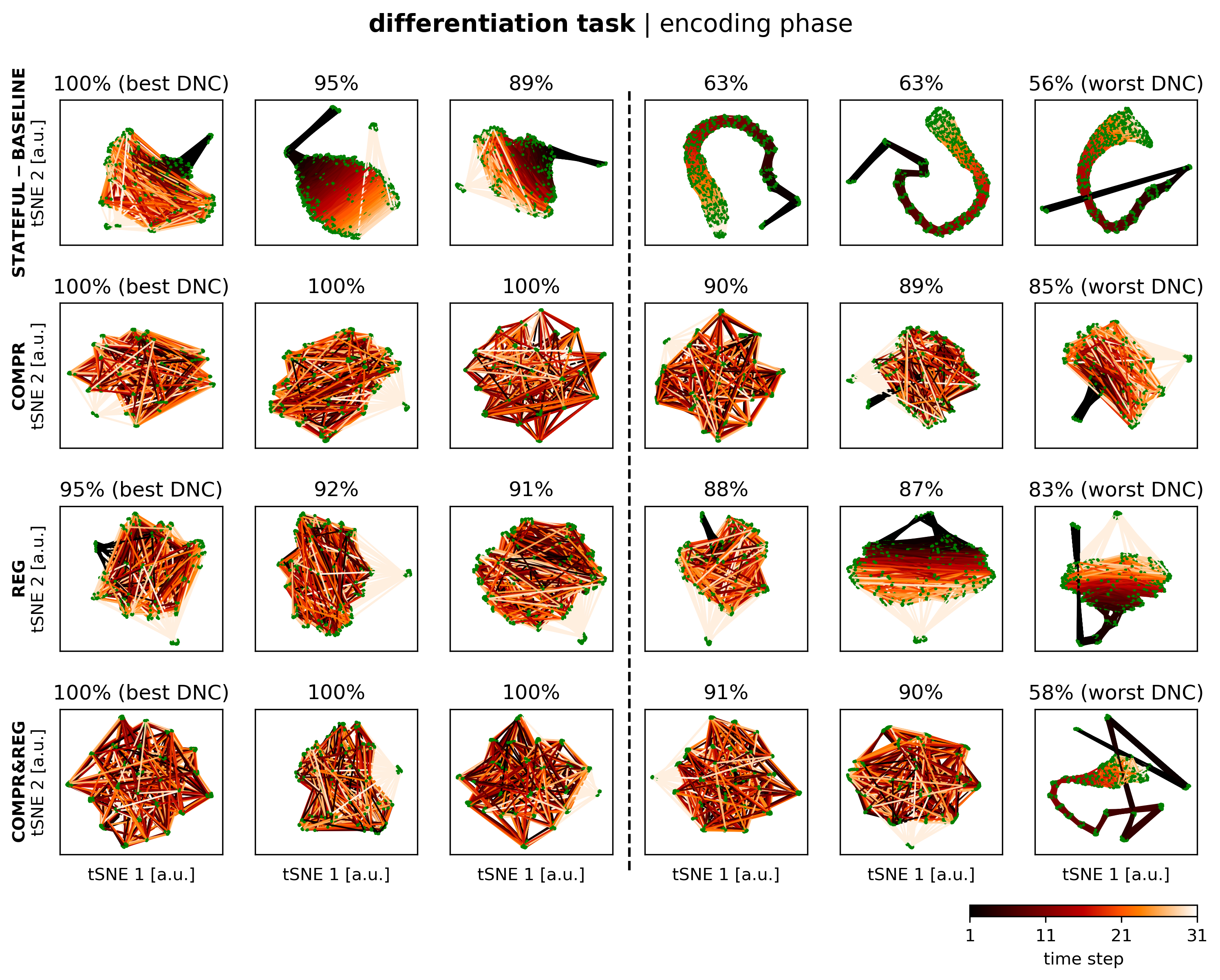}
	\caption{Cell state trajectories from the encoding phase of the differentation task projected with t-SNE.}
	\label{fig:controller_states_differentiation_encoding}
\end{figure*}

\begin{figure*}[!h]
	\centering
	\includegraphics[width=0.7\textwidth]{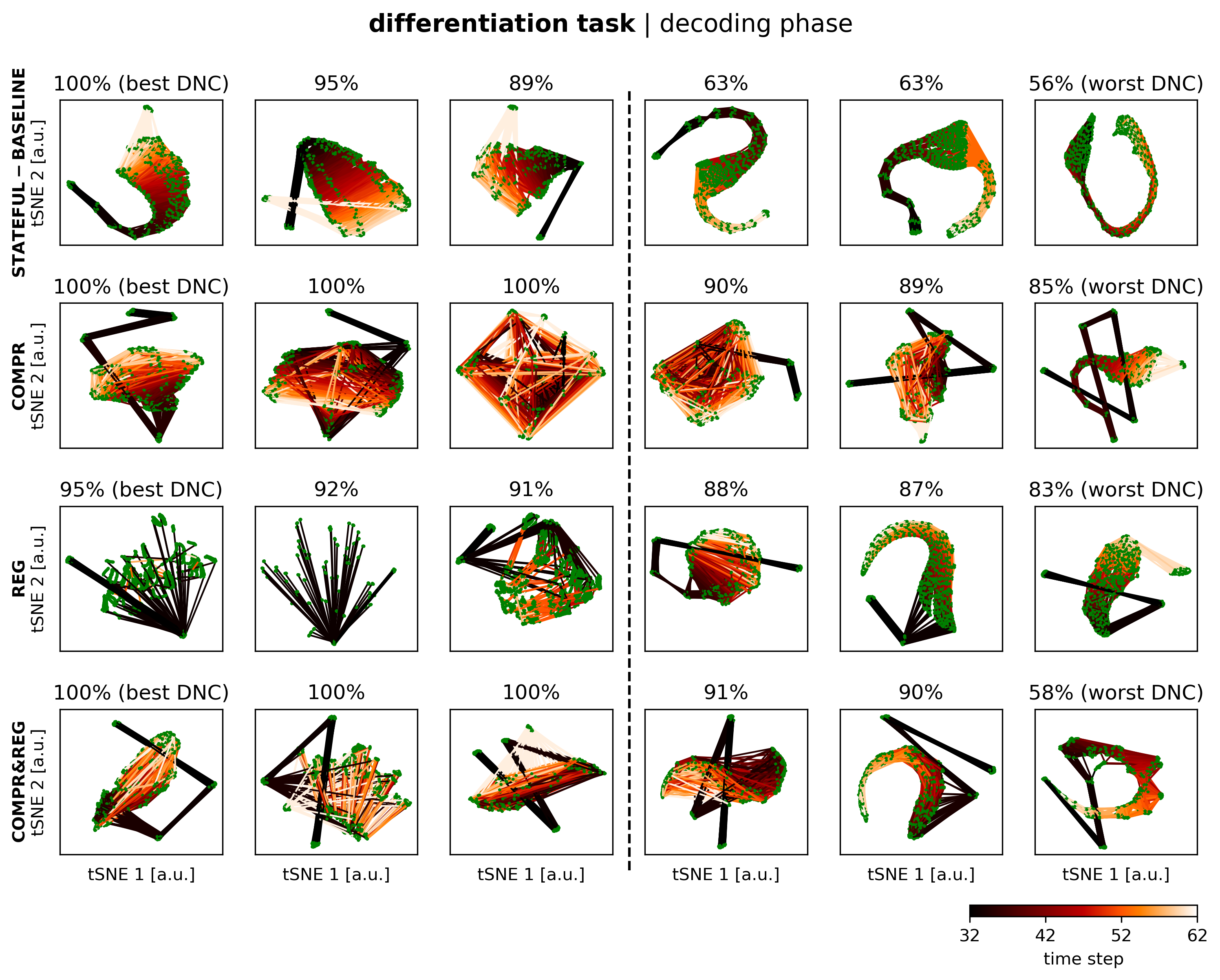}
	\caption{Cell state trajectories from the decoding phase of the differentiation task projected with t-SNE.}
	\label{fig:controller_states_differentiation_decoding}
\end{figure*}

\begin{figure*}[!h]
	\centering
	\includegraphics[width=0.7\textwidth]{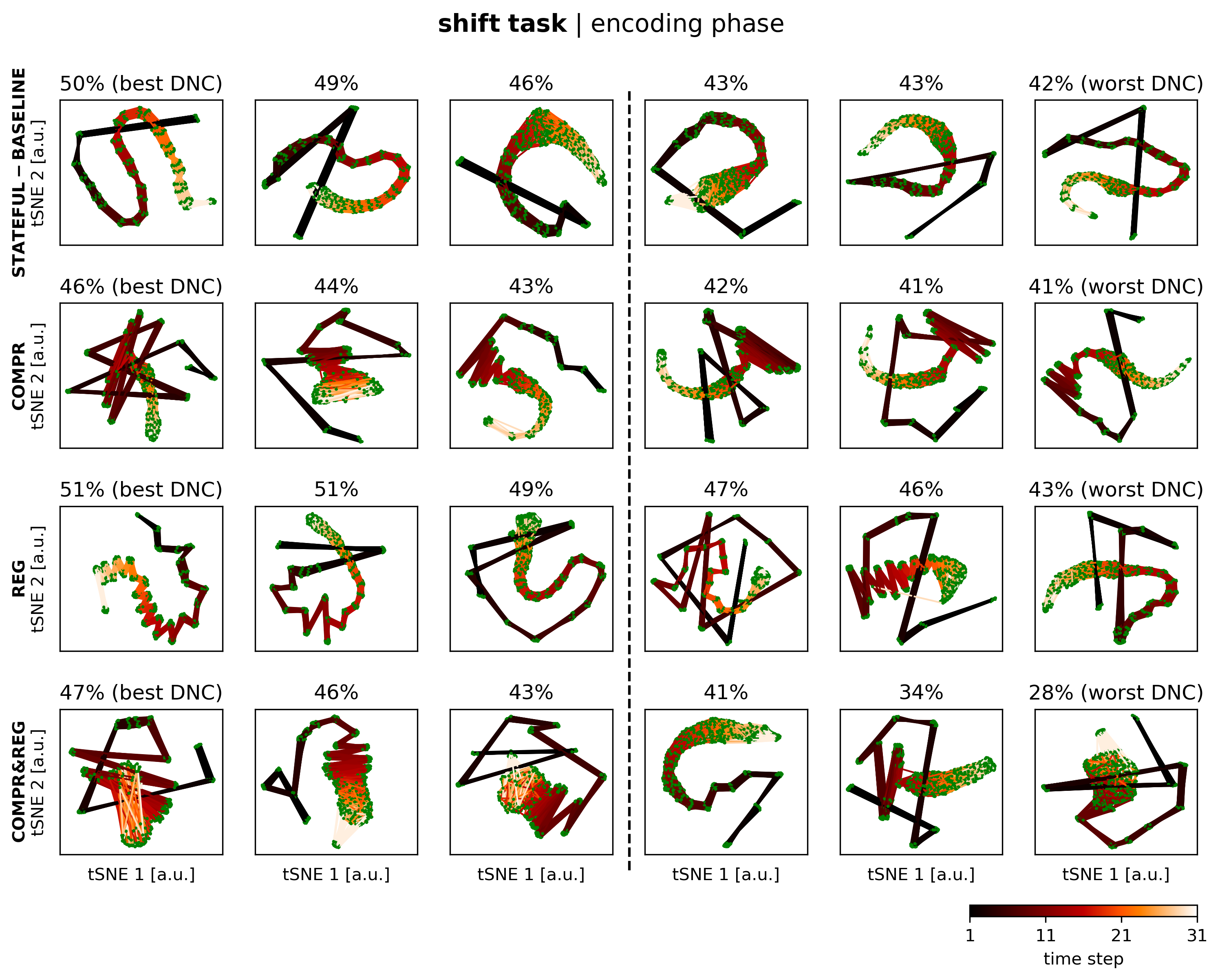}
	\caption{Cell state trajectories from the encoding phase of the shift task projected with t-SNE.}
	\label{fig:controller_states_shift_encoding}
\end{figure*}

\begin{figure*}[!h]
	\centering
	\includegraphics[width=0.7\textwidth]{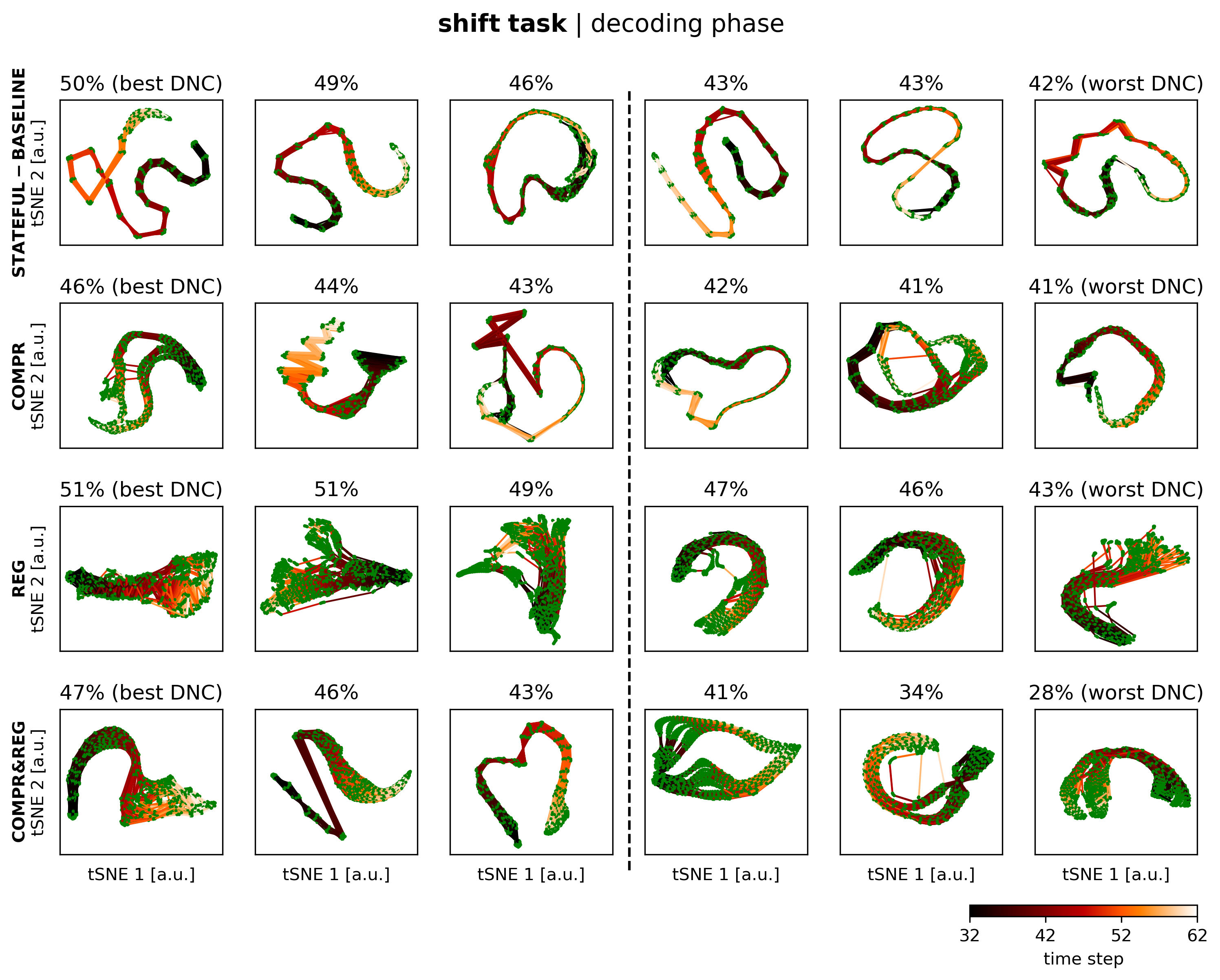}
	\caption{Cell state trajectories from the decoding phase of the shift task projected with t-SNE.}
	\label{fig:controller_states_shift_decoding}
\end{figure*}

\begin{figure*}[!h]
	\centering
	\includegraphics[width=0.7\textwidth]{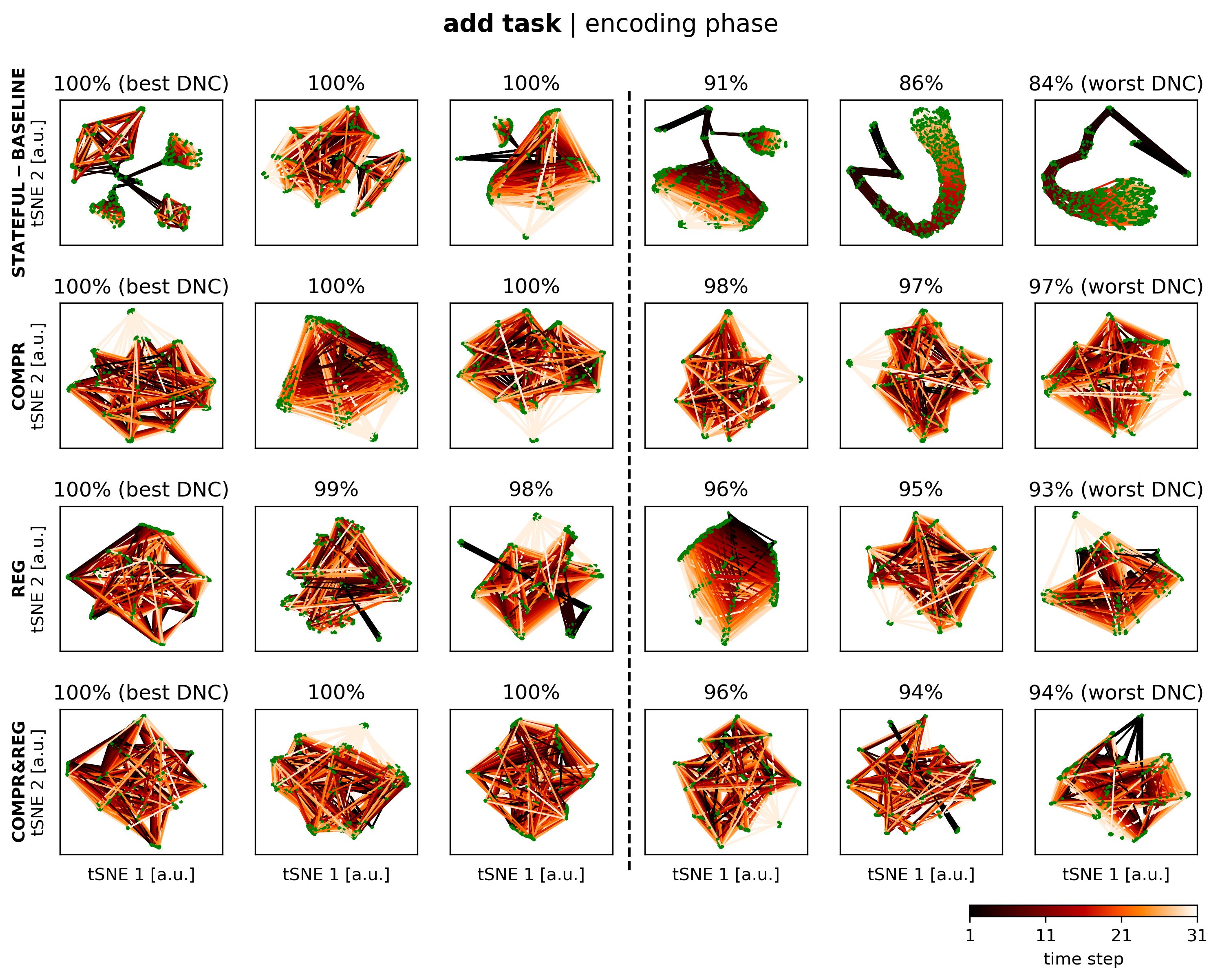}
	\caption{Cell state trajectories from the encoding phase of the add task projected with t-SNE.}
	\label{fig:controller_states_add_encoding}
\end{figure*}

\begin{figure*}[!h]
	\centering
	\includegraphics[width=0.7\textwidth]{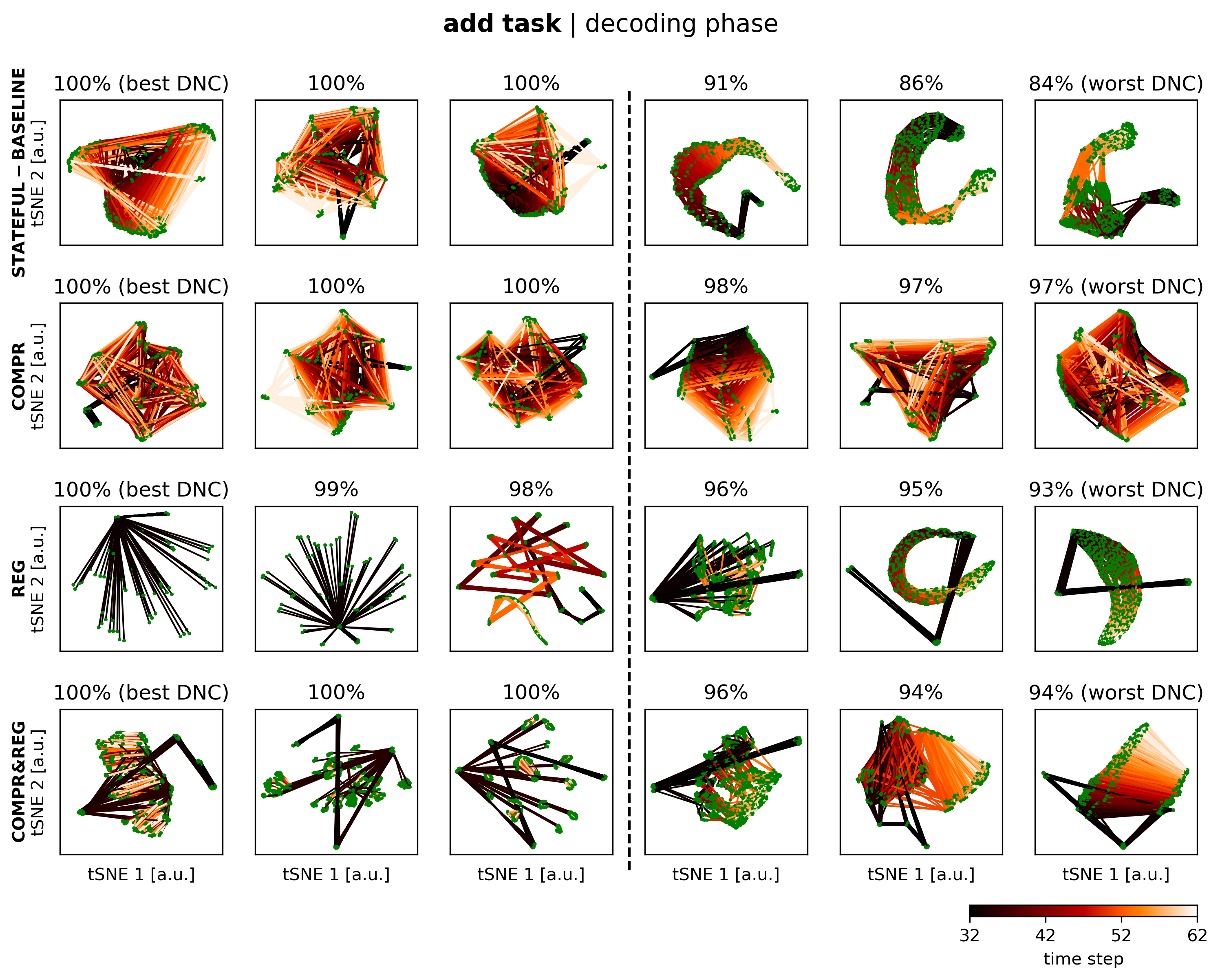}
	\caption{Cell state trajectories from the decoding phase of the add task projected with t-SNE.}
	\label{fig:controller_states_add_decoding}
\end{figure*}

\begin{figure*}[!h]
	\centering
	\includegraphics[width=0.7\textwidth]{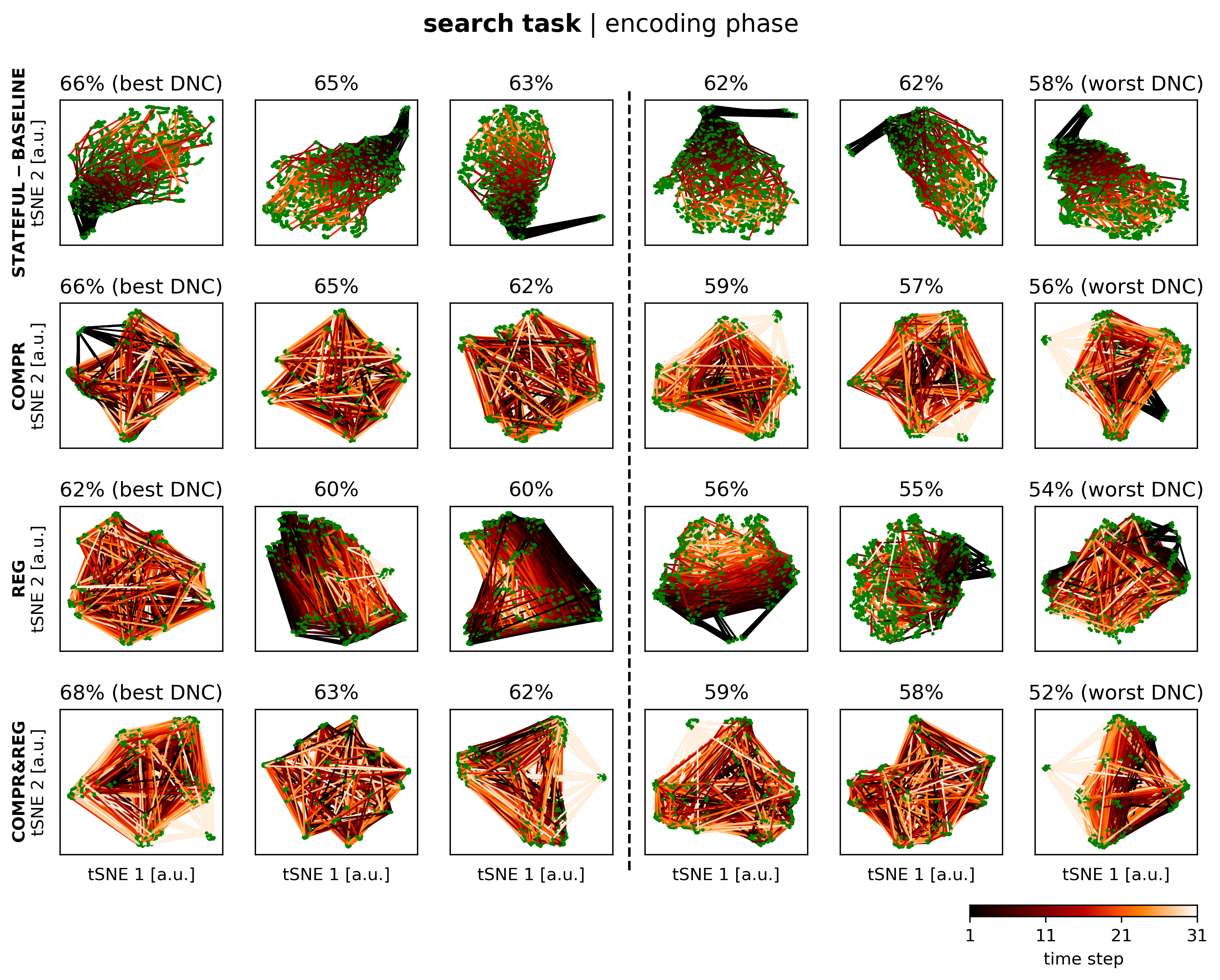}
	\caption{Cell state trajectories from the encoding phase of the search task projected with t-SNE.}
	\label{fig:controller_states_search_encoding}
\end{figure*}

\begin{figure*}[!h]
	\centering
	\includegraphics[width=0.7\textwidth]{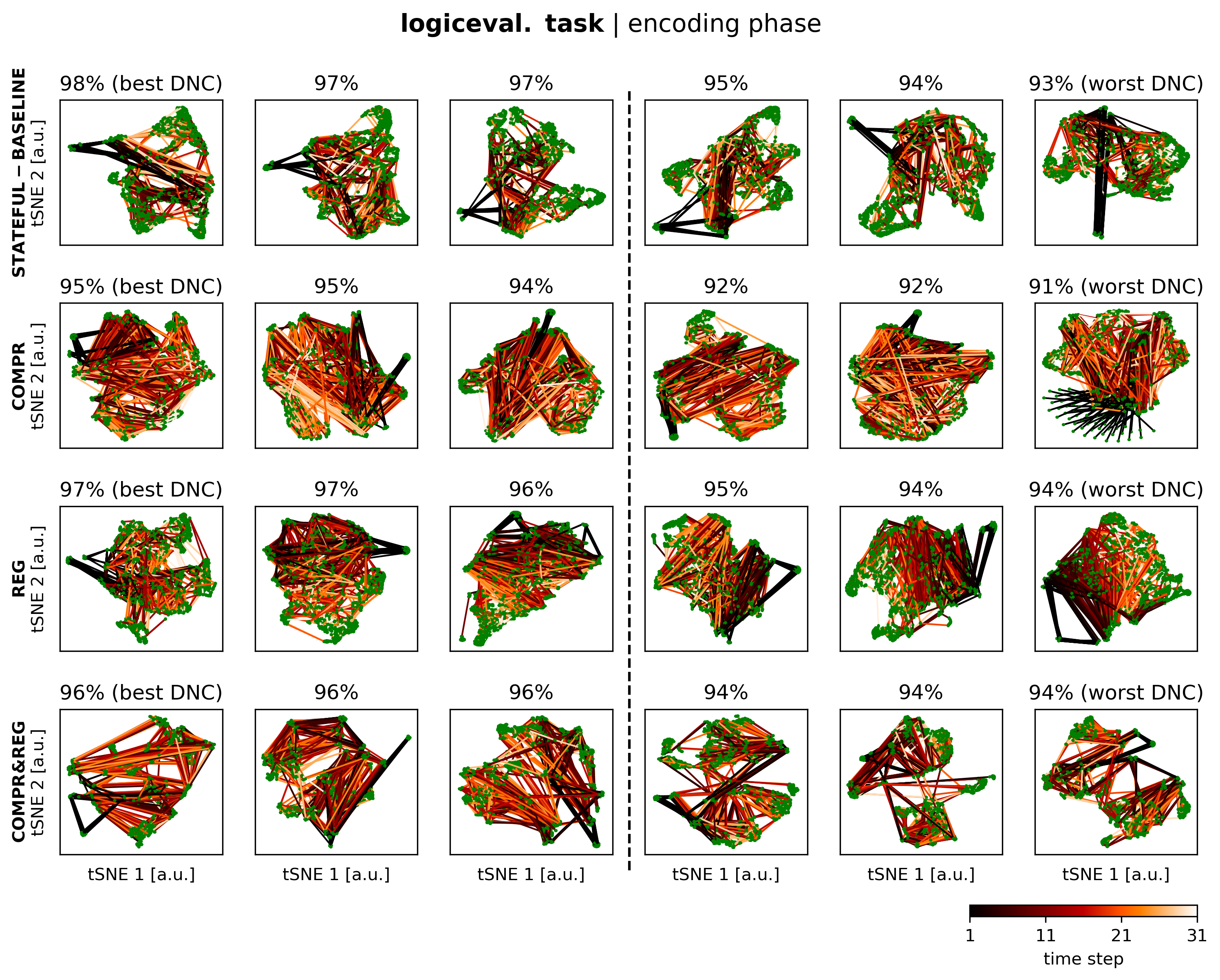}
	\caption{Cell state trajectories from the encoding phase of the logic evaluation task projected with t-SNE.}
	\label{fig:controller_states_logic_encoding}
\end{figure*}

Figures \ref{fig:controller_states_sort_decoding} to \ref{fig:controller_states_logic_encoding} show the remaining state-space trajectory plots. The decoding phase was omitted for the logic evaluation and search tasks. The former had inconsistent decoding phase lengths between samples, and the latter comprised only one time step. The trajectories show the processing of an input sequence with length 30 and the end-of-input or end-of-output flag for the encoding or decoding phase, respectively.

%% file: tables/scores_method_stat_extmem.tex
\begin{tabular}{l|cccc|cccc|cccc|cccc|cccc}
\hline
{} & \multicolumn{4}{|c}{STATEFUL-BASELINE} & \multicolumn{4}{|c}{STATELESS-BASELINE} & \multicolumn{4}{|c}{COMPR} & \multicolumn{4}{|c}{REG} & \multicolumn{4}{|c}{COMPR\&REG} \\
{} &            median & max &    mean & SD &             median & max & mean & SD &  median & max &    mean & SD & median & max &    mean & SD &    median & max &    mean & SD \\
task                  &                   &     &         &    &                    &     &      &    &         &     &         &    &        &     &         &    &           &     &         &    \\
\hline
sort (acc)            &                42 &  48 &      41 &  6 &                 66 &  91 &   68 & 17 & \bf 93 &  96 & \it 89 & 10 &     74 &  98 &      68 & 23 &        91 &  96 &      88 & 10 \\
copy (acc)            &                48 &  89 &      51 & 20 &                 21 &  79 &   30 & 18 &      69 &  91 &      70 & 16 &     82 &  94 &      79 & 12 &   \bf 89 &  96 & \it 86 & 11 \\
differentiation (acc) &                52 &  99 &      59 & 21 &                 26 &  31 &   27 &  3 &      88 &  98 &      79 & 18 &     90 &  94 &      80 & 16 &   \bf 94 &  97 & \it 83 & 23 \\
shift (acc)           &           \bf 30 &  32 &      30 &  2 &                 20 &  20 &   20 &  0 &      28 &  30 &      29 &  2 &     30 &  32 & \it 30 &  2 &        28 &  30 &      27 &  2 \\
add (acc)             &                85 &  99 &      85 & 11 &                 64 &  77 &   64 & 10 &      96 &  99 &      96 &  2 &     93 & 100 &      93 &  7 &   \bf 98 & 100 & \it 98 &  2 \\
logic eval. (acc)     &           \bf 75 &  78 & \it 74 &  2 &                 54 &  56 &   53 &  4 &      70 &  71 &      70 &  1 &     74 &  78 &      73 &  3 &        73 &  75 &      72 &  3 \\
search (hit rate)     &           \bf 19 &  23 & \it 18 &  3 &                 12 &  43 &   16 & 12 &      15 &  26 &      16 &  4 &     15 &  24 &      15 &  4 &        11 &  16 &      12 &  3 \\
\hline
\end{tabular}

%% file: tables/scores_method_test_extmem.tex
\begin{tabular}{l|cccc|ccc|cc|c}
\hline
{} & \multicolumn{4}{c|}{STATEFUL-BASELINE vs} & \multicolumn{3}{c|}{STATELESS-BASELINE vs} & \multicolumn{2}{c|}{COMPR vs} &     REG vs \\
{} &   STATELESS-BASELINE &      COMPR &        REG &  COMPR\&REG &                 COMPR &        REG &  COMPR\&REG &      REG & COMPR\&REG &  COMPR\&REG \\
task            &                      &            &            &            &                       &            &            &          &           &            \\
\hline
sort            &           \bf 0.008 & \bf 0.001 &      0.077 & \bf 0.001 &            \bf 0.018 &      0.671 & \bf 0.022 &    0.059 &     1.000 &      0.088 \\
copy            &           \bf 0.018 &      0.143 & \bf 0.018 & \bf 0.007 &            \bf 0.003 & \bf 0.002 & \bf 0.001 &    0.449 &     0.077 &      0.127 \\
differentiation &           \bf 0.001 &      0.077 &      0.053 &      0.102 &            \bf 0.001 & \bf 0.001 & \bf 0.002 &    0.815 &     0.767 &      0.189 \\
shift           &           \bf 0.001 &      0.127 &      0.767 & \bf 0.026 &            \bf 0.001 & \bf 0.001 & \bf 0.001 &    0.118 &     0.261 & \bf 0.026 \\
add             &           \bf 0.008 &      0.127 &      0.232 & \bf 0.031 &            \bf 0.001 & \bf 0.001 & \bf 0.001 &    0.324 &     0.232 &      0.069 \\
logic eval.     &           \bf 0.001 & \bf 0.001 &      0.671 &      0.143 &            \bf 0.001 & \bf 0.001 & \bf 0.001 &    0.059 &     0.053 &      0.232 \\
search          &                0.291 &      0.127 &      0.059 & \bf 0.004 &                 0.542 &      0.588 &      0.984 &    0.671 &     0.088 &      0.214 \\
\hline
\end{tabular}